\documentclass{article}
\usepackage{svg}
\usepackage{graphicx} 
\usepackage{tikz}
\usepackage{caption}

\usepackage{fontawesome5}

\usepackage{PRIMEarxiv}
\usepackage[numbers]{natbib}
\usepackage{amsmath}
\usepackage{comment}
\usepackage[utf8]{inputenc} 
\usepackage[T1]{fontenc}    
\usepackage{hyperref}       
\usepackage{url}            
\usepackage{booktabs}       
\usepackage{amsfonts}       
\usepackage{nicefrac}       
\usepackage{microtype}      
\usepackage{lipsum}
\usepackage{fancyhdr}       
\usepackage{graphicx}       
\graphicspath{{media/}}     
\usepackage{orcidlink} 

\usepackage{cleveref}
\title{OmniRad: A Radiological Foundation Model for Multi-Task Medical Image Analysis}

\author{
  Luca Zedda\,\orcidlink{0009-0001-8488-1612}, 
  Andrea Loddo\,\orcidlink{0000-0002-6571-3816}, 
  Cecilia Di Ruberto\,\orcidlink{0000-0003-4641-0307}\\
  \\
  \textit{Department of Mathematics and Computer Science, University of Cagliari, Cagliari, Italy} \\
  \texttt{\{luca.zedda,andrea.loddo,cecilia.dir\}@unica.it} \\
}

\begin{document}
\maketitle

\begin{abstract}
Radiological analysis increasingly benefits from pretrained visual representations that can support heterogeneous downstream tasks across imaging modalities. In this work, we introduce OmniRad, a self-supervised radiological foundation model pretrained on 1.2 million medical images, designed with radiology-inspired principles emphasizing representation reuse and cross-task transferability. We evaluate the pretrained encoder under multiple downstream adaptation regimes, including lightweight task-specific adapters with a frozen backbone as well as full end-to-end fine-tuning for classification, allowing us to assess both representation quality and task-specific performance. OmniRad is evaluated on a broad suite of public benchmarks spanning classification and segmentation across multiple modalities. On the MedMNISTv2 collection, OmniRad improves classification F1 by up to 2.05\% over competing foundation models. For dense prediction, OmniRad attains mean Dice score improvements across six MedSegBench datasets when using frozen representations. Qualitative analyses and latent-space visualizations suggest improved feature clustering and modality-related separation. 
\end{abstract}
\begin{center}

\vspace{0.5em}
\noindent
\faGithub\ Code: \url{https://github.com/unica-visual-intelligence-lab/OmniRad}
\quad\\
\faRobot\ Models: \url{https://huggingface.co/collections/Snarcy/omnirad}

\end{center}

\section{Introduction}
\label{sec:intro}

Radiological imaging is fundamental to clinical decision-making, yet image interpretation remains challenging due to inter- and intra-observer variability, subtle pathological patterns, and the scale and heterogeneity of contemporary imaging data. These challenges motivate computational methods that extract quantitative and reproducible information from radiological images.

Radiomics~\cite{lambin_radiomics_2012} addresses this need by transforming medical images into structured quantitative descriptors encoding intensity, morphological, and textural characteristics, typically extracted from segmented regions of interest. While these handcrafted features are interpretable and clinically meaningful, classical radiomics pipelines are highly sensitive to image acquisition protocols, preprocessing strategies, and segmentation quality, limiting reproducibility and scalability in multi-center and multi-modality settings~\cite{zhong_multi-modal_2026}.

Deep learning enables data-driven feature learning directly from images, allowing convolutional and transformer-based models to capture complex anatomical structures and subtle pathological patterns.
However, most deep radiomics models are trained in a fully supervised and task-specific manner, 
requiring large annotated datasets and limiting robustness when transferred across modalities, organs, or institutions.

Self-supervised learning has enabled radiological foundation models pretrained on large collections of unlabeled images~\cite{dantonoli_foundation_nodate}. These models provide transferable representations for downstream tasks such as classification and segmentation, with emerging extensions to vision--language modeling~\cite{zhong_abn-blip_2026}. Despite their promise, existing radiological foundation models often lack sufficient anatomical or modality diversity, and downstream adaptation is typically performed independently for each task, resulting in fragmented feature spaces.
\captionsetup{hypcap=false}
\begin{center}
    \includegraphics[width=0.5\linewidth]{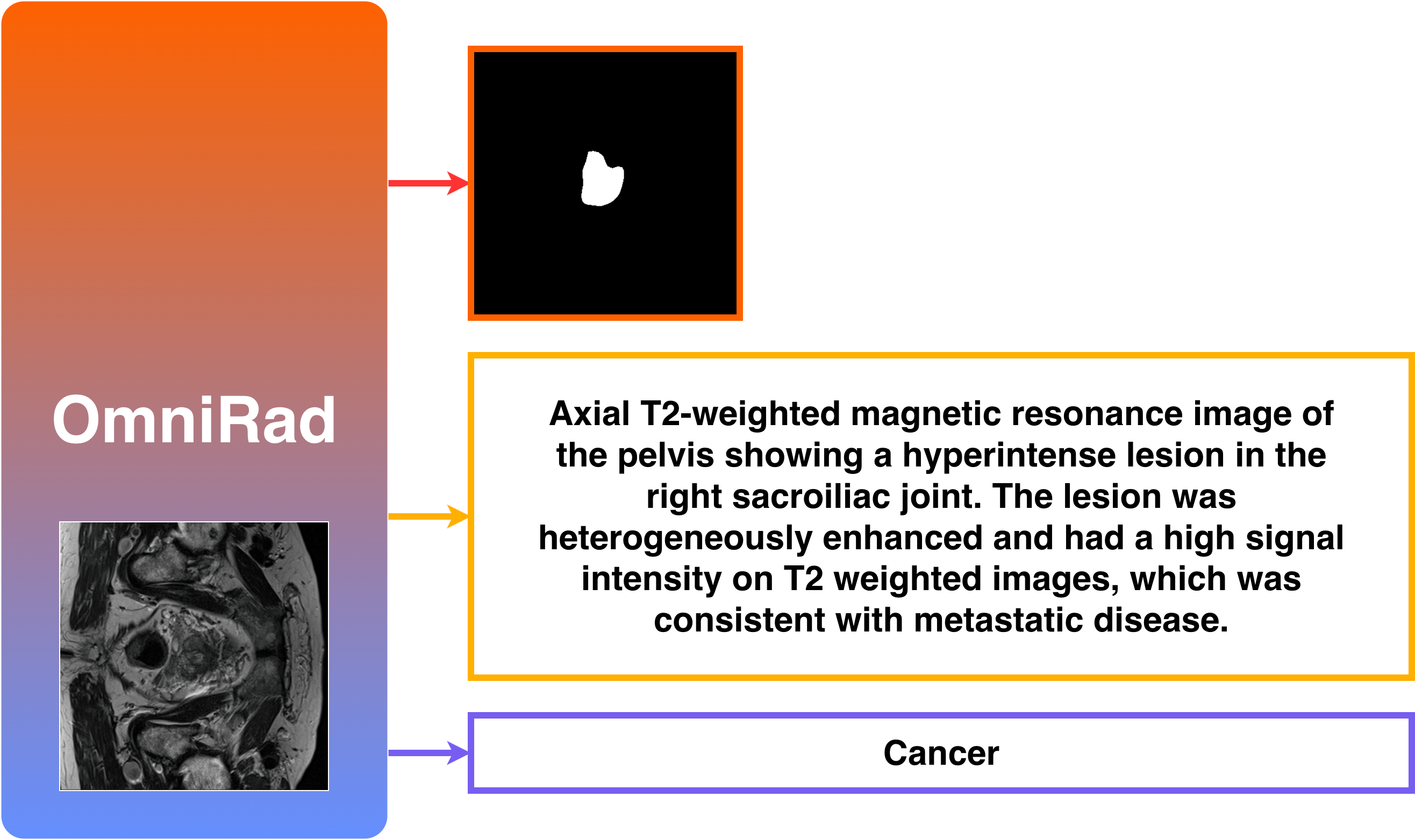}

    \captionof{figure}{Schematic of the proposed OmniRad model. The framework supports multiple tasks by leveraging a shared, unified radiological image foundation model.}
    \label{fig:schema_overview_short}
\end{center}
\captionsetup{hypcap=true}
In clinical practice, radiological workflows encompass multiple related tasks, including diagnostic classification, anatomical or pathological delineation~\cite{okolo_cln_2025}, and, in exploratory research settings, narrative report generation. Although these tasks are commonly addressed independently during model development, they rely on shared visual evidence and complementary information. From a radiomics perspective, the use of inconsistent or task-specific representations undermines feature reuse and limits the reliability of downstream analysis, particularly in longitudinal studies or heterogeneous clinical environments.

To address these limitations, we introduce OmniRad, a radiological foundation model designed to learn stable and transferable radiological representations. OmniRad is pretrained using a self-supervised strategy on large-scale, heterogeneous radiological data spanning multiple modalities and anatomical regions. A single shared encoder is pretrained once and subsequently adapted separately to classification and segmentation tasks, and additionally evaluated in an exploratory vision–language setting, enabling consistent representation reuse across diverse clinical objectives without relying on joint multi-task optimization.

The main contributions of this work are as follows:
\begin{itemize}
    \item We propose OmniRad, a radiological foundation model pretrained on heterogeneous radiological corpora using a self-supervised learning framework.
    \item We introduce a task-agnostic representation paradigm in which a single pretrained encoder is consistently adapted to multiple radiological analysis tasks, promoting feature stability and reuse.
    \item We demonstrate consistent improvements over state-of-the-art medical foundation models across heterogeneous radiological benchmarks, highlighting the robustness and transferability of OmniRad representations.
    \item We present an exploratory proof of concept showing that the learned representations can be extended to vision-language applications without modifying the pretrained encoder.

\end{itemize}
\section{Related Works}
\label{sec:sota}

Classical radiomics established the foundations of quantitative radiological image analysis by transforming medical images into structured and reproducible descriptors. Handcrafted features encoding intensity distributions, morphological properties, and texture patterns extracted from segmented regions of interest have been extensively used for prognosis estimation, treatment response assessment, and disease characterization. A key strength of classical radiomics lies in its interpretability and alignment with clinical reasoning, as features are explicitly defined and can often be linked to known imaging biomarkers~\cite{riberdy_radiomics_2025}.

However, classical radiomics pipelines are highly sensitive to upstream design choices. Variations in image acquisition protocols, reconstruction parameters, preprocessing strategies, and segmentation quality can substantially affect feature distributions, limiting reproducibility across centers and modalities. Moreover, extending handcrafted pipelines to new organs, imaging modalities, or clinical tasks requires significant manual engineering, which constrains scalability and hinders deployment in heterogeneous radiological settings~\cite{fernandez-miranda_retrospective_2024}.

\subsection{Deep Learning-based Radiomics}

Deep learning has expanded the scope of radiomics by enabling data-driven feature learning directly from images. Convolutional and transformer-based architectures can encode complex anatomical structures and subtle pathological patterns that are difficult to capture with handcrafted descriptors~\cite{yang_ddkg_2025}. These learned representations have demonstrated strong performance across a wide range of supervised radiological tasks, including disease classification and organ segmentation~\cite{kim_communication_2025}.

Despite these advances, deep radiomics models often depart from core radiomics principles. Task-specific supervised training on limited annotated datasets can lead to representations that are tightly coupled to particular datasets or imaging protocols, reducing stability and transferability. As a result, features learned for one task or modality may not generalize reliably to other clinical scenarios, limiting their reuse in multi-task or longitudinal radiological workflows.

\subsection{Medical Foundation Models for Radiology}

Self-supervised learning has enabled the emergence of medical foundation models trained on large collections of unlabeled radiological images~\cite{park_self-supervised_2024}. By leveraging generic pretraining objectives, these models reduce reliance on expert annotations and provide transferable visual representations for downstream tasks such as classification and segmentation. Vision transformer backbones pretrained with modern self-supervised strategies have demonstrated particularly strong performance across radiological benchmarks~\cite{zedda_radio_2025}.

However, existing radiological foundation models present several limitations when viewed through a radiomics lens. Pretraining datasets are often restricted to specific modalities or anatomical regions, biasing learned representations and limiting generalization. Downstream adaptation is commonly performed independently for each task, resulting in fragmented feature spaces and inconsistent representations across classification and segmentation~\cite{koleilat_medclip-samv2_2025}. Properties such as feature stability under domain shifts, reproducibility across tasks, and representation reuse are rarely evaluated or explicitly encouraged during model design.

\subsection{Multi-task Applicability in Radiological Workflows}

Radiological practice inherently involves multiple interdependent visual analysis tasks. Diagnostic workflows rely on both global image interpretation and localized anatomical or pathological delineation, requiring representations that capture shared semantic structure across classification and segmentation. While textual reporting provides complementary contextualization~\cite{lang_dacg_2025}, visual analysis tasks form the core of quantitative radiological pipelines.

Most existing approaches address classification and segmentation through task-specific encoders or independently pretrained models, leading to heterogeneous feature spaces across applications. Although this paradigm can yield strong performance on individual benchmarks, it limits the reuse of learned representations and reduces consistency across tasks~\cite{wu_biologically_2024,zhao_multi-task_2023}. From a radiomics perspective, such fragmentation undermines feature stability and complicates downstream analysis, particularly in longitudinal or multi-center settings where consistent representations are essential.

Recent foundation models aim to mitigate these limitations by learning general-purpose visual representations that can be adapted to multiple radiological tasks. However, downstream adaptation is typically performed separately for classification~\cite{zhu_classification_2024,manigrasso_mammography_2025} and segmentation~\cite{gu_segmentanybone_2025}, without mechanisms to explicitly encourage representation consistency across tasks. As a result, the extent to which learned features remain stable and reusable across diverse radiological objectives remains an open question.

These observations motivate approaches that emphasize a shared, task-agnostic radiological encoder, pretrained once and consistently reused across heterogeneous visual analysis tasks~\cite{niu_medical_2025}. Such a paradigm aligns closely with radiomics principles by prioritizing representation stability and transferability, while avoiding the complexity and potential optimization conflicts associated with joint multi-task training.

\subsection{Open Challenges}

The evolution from handcrafted radiomics to deep learning and foundation models has significantly advanced quantitative radiological analysis. However, several challenges remain unresolved. Current approaches often lack sufficient modality and anatomical diversity during pretraining, rely on task-specific adaptation strategies, and do not explicitly enforce radiomics-oriented properties such as feature stability and representation reuse. Moreover, the dominant focus on isolated downstream tasks underutilizes shared structure within radiological data~\cite{shen_multi-modal_2024}.

These limitations highlight the need for unified radiological foundation models that integrate radiomics principles with self-supervised learning and support coherent multi-task adaptation. Such models are essential for developing robust, transferable, and clinically reliable visual representations suitable for real-world radiological workflows~\cite{yao_deep_2025}. OmniRad is proposed to mitigate these critical open challenges.

\section{Method}
\label{sec:method}

The proposed approach for image–based radiological tasks is outlined in this section. In \cref{subsec:proposed_pretrain}, we introduce the radiological pretraining strategy adopted to develop the OmniRad Image Encoder. In \cref{subsec:proposed_cls}, we describe the adaptation of the encoder to global classification tasks. In \cref{subsec:proposed_dense}, we present the extension of OmniRad to dense prediction tasks, with a focus on radiological segmentation. Finally, in \cref{subsec:proposed_captioning}, we provide an exploratory evaluation of the learned representations by employing OmniRad as a frozen visual encoder in a proof-of-concept radiological image captioning setting.

\begin{figure*}
    \centering
    \includegraphics[width=0.8\linewidth]{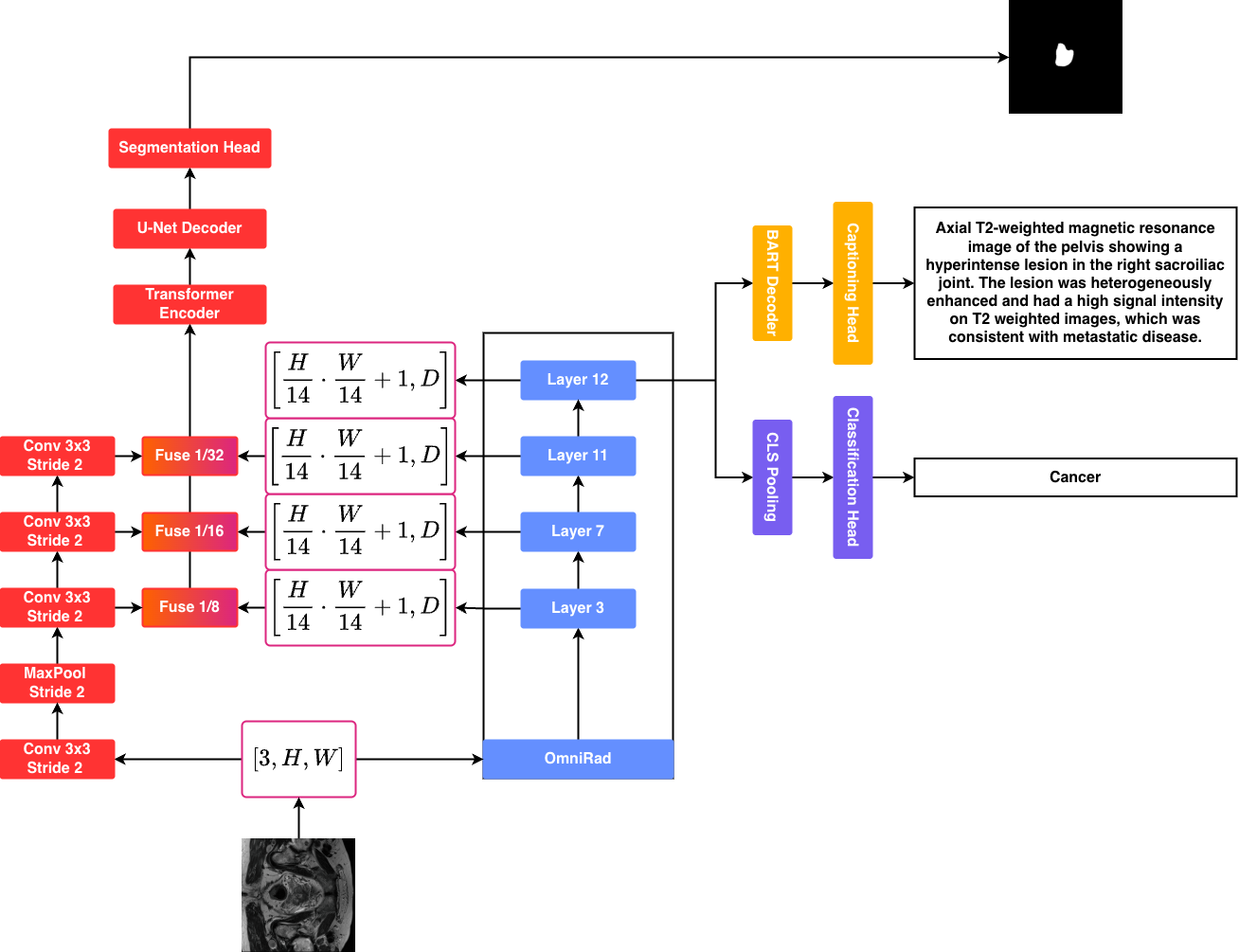}
    \caption{
        Overview of the proposed OmniRad framework for radiological analysis tasks. The OmniRad Image Encoder is shown in blue, the classification branch in purple, the segmentation modules in red, and an exploratory captioning branch in yellow. Feature map dimensionalities are highlighted in pink to illustrate the intermediate representations.
    }
    \label{fig:full_architecture}
\end{figure*}

The overall architecture is illustrated in \cref{fig:full_architecture}. The OmniRad Image Encoder is shown in blue, the classification branch in purple, the segmentation modules in red, the feature map dimensionalities in pink, and the exploratory captioning branch in yellow.

\subsection{OmniRad Pretraining}
\label{subsec:proposed_pretrain}

OmniRad acts as a strong visual encoder for radiological images, primarily due to the adopted self–supervised pretraining strategy. Previous studies \cite{zedda_radio_2025} explored the application of DINO \cite{caron_emerging_2021} and DINOv2 \cite{oquab_dinov2_nodate} for radiological pretraining. However, the direct use of DINOv2 remained an open problem due to suboptimal downstream performance.

OmniRad is pretrained on the RadImageNet\cite{noauthor_radimagenet_nodate} dataset, which provides large–scale and diverse radiological imagery covering multiple anatomical regions and imaging modalities, while providing a clean and structured data source for high-quality representation learning. We further modify the original DINOv2 formulation by retaining only global crops and completely removing local crops, following the design principles introduced in \cite{koch_dinobloom_2024}. This design choice significantly improves training stability and downstream transferability, improving upon \cite{zedda_radio_2025} and speedups approximately by $\times2$ the whole training process due to less required computation.

The training procedure exhibited improved stability and did not require model rollbacks or interventions to mitigate feature collapse, which remains a common issue in self–supervised learning frameworks. Both the small and basevariants are trained using a shared hyperparameter configuration, based on ViT-S and ViT-B backbones, respectively \cite{dosovitskiy_image_nodate}. The complete set of pretraining hyperparameters is reported in \cref{tab:hyperDINO}.

\begin{table}[t!]
\centering
\resizebox{0.5\linewidth}{!}{
\begin{tabular}{lcc}
\toprule
\textbf{Hyperparameter} & \textbf{small models} & \textbf{Base models} \\
\midrule
Student architecture          & VIT small & VIT base \\
Batch size per GPU            & 256 & 196 \\
Patch size                    & \multicolumn{2}{c}{14} \\
Drop path rate (student)      & \multicolumn{2}{c}{0.3} \\
Layer scale                   & \multicolumn{2}{c}{$1 \times 10^{-5}$ } \\
Epochs                        & \multicolumn{2}{c}{10} \\
Base learning rate            & \multicolumn{2}{c}{2e$^{-4}$} \\
Min learning rate             & \multicolumn{2}{c}{1e$^{-6}$} \\
Weight decay (start / end)    & \multicolumn{2}{c}{0.04 / 0.2} \\
Optimizer                     & \multicolumn{2}{c}{AdamW} \\
Teacher momentum (start → end)& \multicolumn{2}{c}{0.994 → 1.0} \\
Warmup teacher temperature    & \multicolumn{2}{c}{0.04} \\
Teacher temperature           & \multicolumn{2}{c}{0.07} \\
DINO loss weight              & \multicolumn{2}{c}{1.0} \\
iBOT loss weight              & \multicolumn{2}{c}{1.0} \\
DINO/iBOT prototypes          & \multicolumn{2}{c}{131072} \\
DINO/iBOT bottleneck dim      & \multicolumn{2}{c}{256} \\
DINO/iBOT head layers         & \multicolumn{2}{c}{3} \\
DINO/iBOT head hidden dim     & \multicolumn{2}{c}{2048} \\
Global crop size              & \multicolumn{2}{c}{224} \\
\bottomrule
\end{tabular}
}
\caption{Main training hyperparameters for the small and base OmniRad Models.}
\label{tab:hyperDINO}
\end{table}

\subsection{Training configuration for radiological classification}
\label{subsec:proposed_cls}

All classification experiments are conducted using both the OmniRad small and basemodels to ensure fair and consistent comparisons with baseline methods. The training procedure follows standard supervised learning with cross–entropy loss.

All classification experiments are conducted using input images resized to $224\times224$ pixels and an effective batch size of $128$ per GPU. Training is performed for $40$ epochs using AdamW with a learning rate of $1\times10^{-5}$, weight decay of $0.01$, and linear warmup over the first $10$ epochs. Gradient clipping with a maximum norm of $1.0$ and standard data augmentations, including random flips, rotations, and resized crops, are applied. 

\subsection{Adapting OmniRad for dense tasks}
\label{subsec:proposed_dense}

Vision Transformers are widely recognized as key architectures for pretraining strategies due to their straightforward scalability to billions of parameters \cite{simeoni_dinov3_2025}. Such extreme scaling, however, requires particular care. A major contributor to their usability is the single–scale representation, which enables multiple pretraining approaches such as masked reconstruction objectives \cite{he_masked_2022}, thereby simplifying the pretraining process.

Despite their strong performance on global tasks such as classification, this design becomes limiting for dense prediction tasks, including detection and segmentation, where hierarchical representations are typically employed \cite{isensee_nnu-net_2021, tian_yolov12_2025}. Several adapter–based approaches \cite{chen2022vitadapter, huang_real-time_2025} have therefore been proposed to bridge this gap.

ViTAdapter introduces substantial computational overhead, resulting in GFLOPs that are impractical for high–throughput radiological workflows. In contrast, the approach of \cite{huang_real-time_2025} focuses on efficient dense adaptation through explicitly lightweight designs.

Following this design philosophy, our method incorporates a parallel convolutional branch that enables efficient adaptation to dense radiological segmentation.

The convolutional branch consists of $3\times3$ convolutions with stride $2$, producing a lightweight hierarchical representation ranging from $1/8$ to $1/32$ of the original image resolution. During dense adaptation, the OmniRad Image Encoder is kept frozen, and only the segmentation–specific modules are optimized.

\noindent\textbf{Spatial-aware Token Adapter and Hybrid Encoder.}

Intermediate token features are extracted from transformer layers $\mathcal{L}=\{3,7,11\}$ and reshaped into spatial feature maps. Earlier layers are selected for higher–resolution feature maps due to their stronger local spatial sensitivity \cite{raghu_vision_2021}. Let $\mathbf{T}_\ell \in \mathbb{R}^{B\times N\times D}$ denote the token embeddings at layer $\ell$, which are rearranged into $\mathbf{V}_\ell \in \mathbb{R}^{B\times D\times H/14 \times W/14}$ and aligned to convolutional priors $\mathbf{C}_s$ at scales $s\in\{8,16,32\}$.

\noindent\textbf{Decoder design.}

The decoding pathway constitutes our main architectural contribution for dense radiological segmentation. Two lightweight upsampling blocks are employed, each performing a $2\times$ bilinear interpolation followed by two $3\times3$ convolutional layers with batch normalization and ReLU activations, and concatenating the upsampled features with those from the corresponding finer scale. The decoding process is defined as
\[
\mathbf{U}_{16} = \psi\!\left(\mathrm{Up}(\mathbf{F}_{32}) \oplus \mathbf{F}_{16}\right), \qquad
\mathbf{U}_{8} = \psi\!\left(\mathrm{Up}(\mathbf{U}_{16}) \oplus \mathbf{F}_{8}\right),
\]
followed by a $1\times1$ convolution producing the final segmentation logits at resolution $H/8 \times W/8$.

All segmentation experiments are conducted using both the OmniRad small and base models to ensure consistent and fair comparisons with baseline methods. Input images are resized to $448\times448$ pixels with a patch size of $14$. A lightweight decoder with two upsampling stages produces predictions at $1/8$ resolution. Models are trained for $20$ epochs using AdamW with a learning rate of $1\times10^{-4}$, weight decay of $1\times10^{-4}$, batch size $16$, and mixed–precision training.
The proposed decoder architectures for the small and base configurations include 14.27 M and 69.76 M trainable parameters, respectively.

\subsection{Exploratory adaptation of OmniRad for image captioning}

\label{subsec:proposed_captioning}

We explore the use of OmniRad as a frozen visual encoder in a proof-of-concept radiological image captioning setting, using an image-language modeling formulation based on BART. For all captioning experiments, the OmniRad basemodel is employed to maintain consistency with the hidden dimensionality of the language model. The OmniRad Image Encoder is kept frozen, while the language modeling components are trained end–to–end.

For all captioning experiments, we use only the OmniRad basemodel. This choice is motivated by its maximal feature capacity among our available models, while matching the hidden size of BART-base($D_l=768$). Using BART-large ($D_l=1024$) would have increased model complexity without clear benefits, making it computationally unnecessary for our setup.

\noindent\textbf{Vision–language projection}

Given an input image $\mathbf{I} \in \mathbb{R}^{3\times H\times W}$, the frozen OmniRad backbone produces token embeddings
\[
\mathbf{T} \in \mathbb{R}^{B\times N\times D_v}, \quad N=\left(\frac{H}{14}\right)^2,
\]
with $D_v=768$ for the basevariant. These features are linearly projected into the BART embedding space
\[
\mathbf{Z} = \mathrm{LN}\!\left(W_p \mathbf{T}\right), \qquad W_p \in \mathbb{R}^{D_v \times D_l},
\]
where $D_l=768$ for BART-base.

\noindent\textbf{Merging visual tokens for captioning}

The BART encoder imposes a fixed and moderate input sequence length. To address this implementation constraint of the HuggingFace BART architecture, we introduce a Patch Merger\cite{renggli_learning_2022} module that compresses the projected OmniRad tokens into a compact set of $K$ latent visual tokens, with $K=64$ in all experiments.

The Patch Merger is implemented as a cross-attention pooling mechanism with learnable queries $\mathbf{Q}\in\mathbb{R}^{K\times D_l}$. Given the projected visual tokens $\mathbf{Z}$, the pooled representation is computed as
\[
\mathbf{A} = \mathrm{softmax}\!\left(\frac{\mathbf{Q}\mathbf{Z}^\top}{\sqrt{D_l}}\right), \qquad
\mathbf{P} = \mathbf{A}\mathbf{Z},
\]
yielding $\mathbf{P}\in\mathbb{R}^{B\times K\times D_l}$, which acts as a visual prefix for the language model.

\noindent\textbf{Language decoding}
The pooled visual tokens $\mathbf{P}$ are provided to the BART encoder as input embeddings, while the BART decoder generates captions autoregressively. The training objective corresponds to the standard autoregressive language modeling loss
\[
\mathcal{L}_{\text{cap}} = -\sum_{t} \log p(y_t \mid y_{<t}, \mathbf{P}),
\]
where $y_t$ denotes the caption tokens.
All captioning experiments are trained for $20$ epochs using AdamW with a learning rate of $5\times10^{-5}$ and mixed–precision training. An effective batch size of $64$ is obtained through gradient accumulation. During inference, captions are generated using beam search with $5$ beams and a maximum length of $64$ tokens.
This experiment is intended to assess the semantic alignment of the learned visual representations with radiological language, rather than to provide a clinically validated report generation system.
\section{Experimental Evaluation}
\label{sec:evaluation}

In this section, we present the methodological evaluation of the proposed OmniRad foundation model across all considered tasks. Specifically, in \cref{subsec:cls_results} we report classification results on five radiological benchmark datasets. In \cref{subsec:seg_results}, we present segmentation results obtained by adapting OmniRad to dense prediction tasks, covering both binary and multiclass segmentation across eight benchmark datasets. Finally, in \cref{subsec:captioning_results}, we report an exploratory set of experiments using a BART-based~\cite{lewis_bart_2020} framework to assess the suitability of OmniRad representations for radiological image captioning.
Our experiments were conducted on a workstation with 2 A100 80gb, with 5 iterations for each experiment to produce a fair comparison.
\subsection{Datasets}
\label{subsec:datasets}

The pretraining and evaluation pipeline relies on complementary datasets covering large-scale visual pretraining, image-level classification, dense segmentation, and multimodal image–text learning. This design enables the assessment of representation transfer across heterogeneous medical imaging tasks. For all datasets, we used official train, val, and test splits to encourage comparisons and evaluation fairness. We report a full list and brief description of the used dataset in \cref{tab:datasets_all}.

\textbf{Pretraining dataset.}
Large-scale visual pretraining is performed using RadImageNet~\cite{noauthor_radimagenet_nodate}, a curated collection of radiological images spanning multiple modalities, including CT, MR, and ultrasound. RadImageNet provides over one million images annotated across 165 classes, covering diverse anatomical regions and pathologies. Its scale and heterogeneity make it suitable for learning robust, modality-agnostic visual representations that generalize well to downstream tasks.

\textbf{Classification datasets.}
For image-level classification evaluation, we use representative benchmarks from the MedMNISTv2~\cite{medmnistv2} family and breast ultrasound datasets. Specifically, PneumoniaMNIST and BreastMNIST provide binary classification tasks on chest X-ray and breast ultrasound images, respectively, while OrganAMNIST, OrganCMNIST, and OrganSMNIST offer multi-class organ classification from abdominal CT scans. These datasets feature standardized splits and controlled image resolutions, enabling reliable and reproducible evaluation of classification performance across different anatomical regions and modalities.

\textbf{Segmentation datasets.}
For dense prediction and pixel-level evaluation, we adopt MedSegBench~\cite{kus_medsegbench_2024}, a comprehensive benchmark for medical image segmentation. MedSegBench aggregates 35 datasets comprising more than 60,000 images across diverse modalities such as ultrasound, MRI, CT, X-ray, OCT, dermoscopy, endoscopy, fundus imaging, and microscopy. It includes both binary and multi-class segmentation tasks with up to 19 classes, standardized train, validation, and test splits, and unified preprocessing at multiple spatial resolutions, allowing systematic assessment of segmentation robustness across heterogeneous clinical settings. For our experiments, we selected only the radiology-related datasets that compose the MedSegBench collection.

\textbf{Exploratory captioning and multimodal datasets.}
To evaluate visual–semantic alignment, we use ROCOv2~\cite{ruckert_rocov2_2024}, a large-scale multimodal dataset consisting of radiological images paired with textual captions and medical concepts extracted from the PMC Open Access Subset. ROCOv2 contains nearly 80,000 image–text pairs spanning multiple radiological modalities.

\begin{table*}[h!]
    \centering
    \resizebox{1\textwidth}{!}{%
    \begin{tabular}{lllccc}
    \toprule
    \textbf{Dataset} & \textbf{Modality} & \textbf{Task} & \textbf{Classes}  & \textbf{Total samples} & \textbf{Splits (train/val/test)} \\
    \midrule
    RadImageNet & CT / MR / Ultrasound & Pretrain & 165 & 1,350,000 & 1,080,000 / 135,000 / 135,000\\
    \midrule
    PneumoniaMNIST & Chest X-Ray & Classification & 2 & 5,856 & 4,708 / 524 / 624 \\
    BreastMNIST & Breast Ultrasound & Classification & 2 & 780 & 546 / 78 / 156 \\
    OrganAMNIST & Abdominal CT & Classification & 11 & 58,830 & 34,561 / 6,491 / 17,778 \\
    OrganCMNIST & Abdominal CT & Classification & 11 & 23,583 & 12,975 / 2,392 / 8,216 \\
    OrganSMNIST & Abdominal CT & Classification & 11 & 25,211 & 13,932 / 2,452 / 8,827 \\
    \midrule
    PandentalMSBench & X-Ray & Segmentation & 2 & 116 & 81 / 11 / 24 \\
    Promise12MSBench & MRI Prostate & Segmentation & 2 & 1,473 & 1,031 / 147 / 295 \\
    USforKidneyMSBench & Ultrasound Kidney & Segmentation & 2 & 4,586 & 3,210 / 458 / 918 \\
    UltrasoundNerveMSBench & Ultrasound Neck & Segmentation & 2 & 2,323 & 1,651 / 223 / 449 \\
    BusiMSBench & Ultrasound Breast & Segmentation & 2 & 647 & 452 / 64 / 131 \\
    CovidQUExMSBench & Chest X-Ray & Segmentation & 2 & 2,913 & 1,864 / 466 / 583 \\
    FHPsAOPMSBench & Ultrasound Fetal head, pubic symphysis & Segmentation & 3 & 4,000 & 2,800 / 400 / 800 \\
    MosMedPlusMSBench & CT Lung & Segmentation & 2 & 2,729 & 1,910 / 272 / 547 \\
    \midrule
    ROCOv2 & Multiple & Captioning & N/A & 79,789 & 59,958 / 9,904 / 9,927 \\
    \bottomrule
    \end{tabular}%
    }
    \caption{Overview of datasets used in experiments. For each dataset, we report imaging modality, primary task, number of target classes where applicable, total image count, and the train/validation/test splits used.}
    \label{tab:datasets_all}
\end{table*}

\subsection{Classification Results}
\label{subsec:cls_results}

\begin{table}[h!]
    \centering
    \resizebox{0.6\textwidth}{!}{%
    \begin{tabular}{lccc}
    \toprule
    \textbf{Model} & \textbf{Accuracy (\%)} & \textbf{F1 (\%)} & \textbf{AUC (\%)} \\
    \midrule
    Resnet18~\cite{medmnistv2}& 83.3 & - & 89.1  \\
    Resnet50~\cite{medmnistv2}& 84.2 & - & 86.6 \\
    auto-sklearn~\cite{medmnistv2}& 80.3 & - & 83.6 \\
    AutoKeras~\cite{medmnistv2}& 83.1 & - & 87.1 \\
    Google AutoML Vision~\cite{medmnistv2}& 86.1 & - & 91.9 \\
    DenseNet121~\cite{singh_dynamic_2024} & 86.3 & - & 90.1 \\
    Swin Transformer~\cite{schafer_overcoming_2024} & 86.92 $\pm$ 1.04 & - & 86.26 $\pm$ 1.38 \\
    R-LLM~\cite{Lai_2024_CVPR} & 87.17 & - & 88.23\\
    Med ViT tiny~\cite{manzari_medvit_2023} & 89.6 & - & 93.4\\
    Med ViT small~\cite{manzari_medvit_2023} & 89.7 & - & 93.8\\
    Med ViT base~\cite{manzari_medvit_2023} & 88.3 & - & 92.9\\\hline
    DINO small & 74.57 $\pm$ 4.17 & 59.14 $\pm$ 7.56 & 71.78 $\pm$ 4.41 \\
    DINO base & 83.12 $\pm$ 0.98 & 75.48 $\pm$ 1.77 & 83.16 $\pm$ 1.05 \\
    DINOv2 small & 73.50 $\pm$ 2.67 & 50.63 $\pm$ 6.57 & 67.28 $\pm$ 7.19 \\
    DINOv2 base & 69.02 $\pm$ 7.59 & 49.07 $\pm$ 6.35 & 63.22 $\pm$ 13.01 \\
    DINOv3 small & 88.03 $\pm$ 2.06 & 83.82 $\pm$ 3.05 & 89.28 $\pm$ 2.49 \\
    DINOv3 base & 89.74 $\pm$ 3.63 & 87.22 $\pm$ 4.02 & 94.12 $\pm$ 1.26 \\
    \hline
    Radio DINO small  & 91.67 $\pm$ 0.68 & 88.98 $\pm$ 1.0 & 95.55 $\pm$ 1.55 \\
    Radio DINO base & 90.38 $\pm$ 1.84 & 87.69 $\pm$ 3.5 & 95.51 $\pm$ 1.7 \\
    \hline
    OmniRad small & \textbf{91.83 $\pm$ 0.32} & \textbf{89.42 $\pm$ 0.18} & \textbf{95.56 $\pm$ 1.00} \\
    OmniRad base & 89.96 $\pm$ 3.29 & 87.32 $\pm$ 3.72 & 93.70 $\pm$ 0.32 \\
    \bottomrule
    \end{tabular}%
    }
    \caption{Performance comparison of different models on the BreastMNIST dataset. The table reports accuracy, F1, and AUC, with an indication of the standard deviation when available. The best results are emphasized in bold.}
    \label{tab:performance_BreastMNIST}
\end{table}

\begin{table}[h!]
    \centering
    \resizebox{0.6\textwidth}{!}{%
    \begin{tabular}{lccc}
    \toprule
    \textbf{Model} & \textbf{Accuracy (\%)} & \textbf{F1 (\%)} & \textbf{AUC (\%)} \\
    \midrule
    Resnet18~\cite{medmnistv2}& 86.4 & - &  95.6 \\
    Resnet50~\cite{medmnistv2}& 88.4 & - &  96.2 \\
    auto-sklearn~\cite{medmnistv2}& 85.5 & - &  94.2 \\
    AutoKeras~\cite{medmnistv2}& 87.8 & - &  94.7 \\
    Google AutoML Vision~\cite{medmnistv2}& 94.6 & - &  99.1 \\
    GCNN-EC~\cite{singh_dynamic_2024} & 87.7 & - & 95.4 \\
    Swin Transformer~\cite{schafer_overcoming_2024} & 91.54 $\pm$ 0.48 & - & 98.22 $\pm$ 0.37 \\
    R-LLM~\cite{Lai_2024_CVPR} & 93.91 & - & 98.01\\
    Med ViT tiny~\cite{manzari_medvit_2023} & 94.9 & - & 99.3\\
    Med ViT small~\cite{manzari_medvit_2023} & \textbf{96.1} & - & 99.5\\
    Med ViT base~\cite{manzari_medvit_2023} & 92.1 & - & 99.1\\
    \hline
    DINO small & 83.71 $\pm$ 1.44 & 80.95 $\pm$ 2.14 & 93.89 $\pm$ 0.79 \\
    DINO base & 90.70 $\pm$ 2.00 & 89.46 $\pm$ 2.41 & 98.52 $\pm$ 1.13 \\
    DINOv2 small & 81.04 $\pm$ 1.73 & 77.75 $\pm$ 2.21 & 92.21 $\pm$ 1.19 \\
    DINOv2 base & 79.49 $\pm$ 0.48 & 75.41 $\pm$ 0.66 & 91.72 $\pm$ 0.13 \\
    DINOv3 small & 93.06 $\pm$ 0.74 & 92.27 $\pm$ 0.86 & 99.26 $\pm$ 0.09 \\
    DINOv3 base & 93.91 $\pm$ 0.68 & 93.26 $\pm$ 0.78 & 98.65 $\pm$ 0.11 \\
    \hline
    Radio DINO small  & 91.83 $\pm$ 1.03 & 90.86 $\pm$ 1.27 & 98.86 $\pm$ 0.23 \\
    Radio DINO base & 93.91 $\pm$ 1.11 & 93.29 $\pm$ 1.31 & 98.93 $\pm$ 0.45 \\
    \hline
    OmniRad small & \textbf{95.30 $\pm$ 1.09} & \textbf{94.85 $\pm$ 1.25} & 99.31 $\pm$ 0.32 \\
    OmniRad base & 94.66 $\pm$ 0.33 & 94.15 $\pm$ 0.39 & \textbf{99.36 $\pm$ 0.12} \\
    \bottomrule
    \end{tabular}%
    }
    \caption{Performance comparison of different models on the PneumoniaMNIST dataset. The table reports accuracy, F1, and AUC, with standard deviation when available. The best results are emphasized in bold.}
    \label{tab:performance_PneumoniaMNIST}
\end{table}

We evaluate OmniRad using Accuracy, F1 score, and AUC, as summarized in~\Cref{tab:performance_BreastMNIST,tab:performance_PneumoniaMNIST,tab:performance_OrganAMNIST,tab:performance_OrganCMNIST,tab:performance_OrganSMNIST}.

\noindent\textbf{Overall performance
}
OmniRad consistently achieves the strongest or near–strongest classification performance across all evaluated datasets. On BreastMNIST, OmniRad small attains the highest F1 score ($89.42\%$), outperforming all convolutional, transformer, and foundation model baselines, including Radio DINO base. On PneumoniaMNIST, OmniRad small establishes the best and F1 score ($94.85\%$), confirming strong generalization on chest radiographs. Based on size only, both datasets are one order of magnitude smaller compared to the other selected for classification, showcasing the high adaptability of OmniRad to limited data sources.

\noindent\textbf{Multi–organ CT benchmarks}  

Across OrganAMNIST, OrganCMNIST, and OrganSMNIST, OmniRad consistently achieves the highest F1 scores, outperforming both Radio DINO and DINOv3. Specifically, on OrganAMNIST, OmniRad small reaches $97.30\%$ F1, exceeding DINOv3 base $97.28\%$  and Radio DINO base $97.20\%$; on OrganCMNIST, OmniRad base attains $95.45\%$ F1, improving over DINOv3 small $94.74\%$  by $+0.71\%$; and on OrganSMNIST, OmniRad base achieves $80.97\%$ F1, surpassing DINOv3 small $78.92\%$ F1 by $+2.05\%$ and Radio DINO base $78.15\%$ F1 by $+2.82\%$. While DINOv3 benefits from pretraining on a much larger, non-radiological corpus, these results show that a specialized foundation model can yield superior performance on medical imaging benchmarks under the same training configuration.

\begin{table}[h!]
    \centering
    \resizebox{0.6\textwidth}{!}{%
    \begin{tabular}{lccc}
    \toprule
    \textbf{Model} & \textbf{Accuracy (\%)} & \textbf{F1 (\%)} & \textbf{AUC (\%)} \\
    \midrule
    Resnet18~\cite{medmnistv2}& 95.1 & - &  99.8 \\
    Resnet50~\cite{medmnistv2}& 94.7 & - &  99.8 \\
    auto-sklearn~\cite{medmnistv2}& 76.2 & - &  96.3 \\
    AutoKeras~\cite{medmnistv2}& 90.5 & - &  99.4 \\
    Google AutoML Vision~\cite{medmnistv2}& 88.6 & - &  99.0 \\
    DenseNet121~\cite{singh_dynamic_2024} & 93.5 & - & 99.7 \\
    R-LLM~\cite{Lai_2024_CVPR} & 95.22 & - & 99.78\\
    Med ViT tiny~\cite{manzari_medvit_2023} & 93.1 & - & 99.5\\
    Med ViT small~\cite{manzari_medvit_2023} & 92.8 & - & 99.6\\
    Med ViT base~\cite{manzari_medvit_2023} & 94.3 & - & 99.7\\\hline
    DINO small & 96.45 $\pm$ 0.58 & 95.93 $\pm$ 0.60 & 99.88 $\pm$ 0.03 \\
    DINO base & 96.79 $\pm$ 0.44 & 96.39 $\pm$ 0.61 & 99.87 $\pm$ 0.04 \\
    DINOv2 small & 95.07 $\pm$ 0.42 & 94.55 $\pm$ 0.41 & 99.83 $\pm$ 0.01 \\
    DINOv2 base & 94.56 $\pm$ 1.13 & 94.19 $\pm$ 1.13 & 99.76 $\pm$ 0.08 \\
    DINOv3 small & 97.06 $\pm$ 0.21 & 96.83 $\pm$ 0.30 & 99.87 $\pm$ 0.02 \\
    DINOv3 base & 97.61 $\pm$ 0.42 & 97.28 $\pm$ 0.47 & 99.88 $\pm$ 0.09 \\
    \hline
    Radio DINO small  & 96.83 $\pm$ 0.37 & 96.47 $\pm$ 0.36 & 99.92 $\pm$ 0.03 \\
    Radio DINO base & 97.35 $\pm$ 0.55 & 97.20 $\pm$ 0.61 & 99.93 $\pm$ 0.03 \\
    \hline
    OmniRad small & \textbf{97.62 $\pm$ 0.39} & \textbf{97.30 $\pm$ 0.32} & \textbf{99.95 $\pm$ 0.05} \\
    OmniRad base & 97.14 $\pm$ 0.51 & 96.98 $\pm$ 0.55 & 99.89 $\pm$ 0.03 \\
    \bottomrule
    \end{tabular}%
    }
    \caption{Performance comparison of different models on the OrganAMNIST dataset. The table reports accuracy, F1, and AUC, with standard deviation when available. The best results are emphasized in bold.}
    \label{tab:performance_OrganAMNIST}
\end{table}

\noindent\textbf{Model capacity effects}

Consistent trends are observed with respect to model scale. The small variant dominates on datasets characterized by limited anatomical diversity and visually localized structures, such as BreastMNIST, PneumoniaMNIST, and OrganAMNIST. In contrast, the base variant consistently outperforms on OrganCMNIST and OrganSMNIST, where higher anatomical heterogeneity and richer class distributions benefit from increased representational capacity.

Globally OmniRad establishes the strongest classification performance across heterogeneous MedMNIST benchmarks, with consistent improvements over existing foundation models and prior domain–specific architectures, confirming the effectiveness of its radiological pretraining strategy for general–purpose radiological recognition.

\begin{table}[h!]
    \centering
    \resizebox{0.6\textwidth}{!}{%
    \begin{tabular}{lccc}
    \toprule
    \textbf{Model} & \textbf{Accuracy (\%)} & \textbf{F1 (\%)} & \textbf{AUC (\%)} \\
    \midrule
    Resnet18~\cite{medmnistv2}& 92.0 & - &  99.4 \\
    Resnet50~\cite{medmnistv2}& 91.1 & - &  99.3 \\
    auto-sklearn~\cite{medmnistv2}& 82.9 & - &  97.6 \\
    AutoKeras~\cite{medmnistv2}& 87.9 & - &  99.0 \\
    Google AutoML Vision~\cite{medmnistv2}& 87.7 & - &  98.8 \\
    EfficientNetB0~\cite{singh_dynamic_2024} & 90.5 & - & 99.2 \\
    Med ViT tiny~\cite{manzari_medvit_2023} & 90.1 & - & 99.1\\
    Med ViT small~\cite{manzari_medvit_2023} & 91.6 & - & 99.3\\
    Med ViT base~\cite{manzari_medvit_2023} & 92.2 & - & 99.4\\\hline
    DINO small & 93.67 $\pm$ 0.26 & 92.91 $\pm$ 0.36 & 99.67 $\pm$ 0.16 \\
    DINO base & 94.28 $\pm$ 1.44 & 94.06 $\pm$ 1.62 & 99.72 $\pm$ 0.05 \\
    DINOv2 small & 91.37 $\pm$ 0.49 & 89.83 $\pm$ 0.72 & 99.38 $\pm$ 0.28 \\
    DINOv2 base & 85.73 $\pm$ 1.36 & 82.61 $\pm$ 2.20 & 98.84 $\pm$ 0.21 \\
    DINOv3 small & 95.31 $\pm$ 0.32 & 94.74 $\pm$ 0.30 & 99.78 $\pm$ 0.03 \\
    DINOv3 base & 95.18 $\pm$ 0.90 & 94.46 $\pm$ 1.10 & 99.79 $\pm$ 0.01 \\
    \hline
    Radio DINO small  & 94.3 $\pm$ 0.31 & 93.63 $\pm$ 0.25 & 99.8 $\pm$ 0.02 \\
    Radio DINO base & 95.11 $\pm$ 0.71 & 94.57 $\pm$ 0.82 & 99.86 $\pm$ 0.04 \\
    \hline
    OmniRad small & 95.49 $\pm$ 0.78 & 94.86 $\pm$ 0.89 & 99.83 $\pm$ 0.03 \\
    OmniRad base & \textbf{96.02 $\pm$ 0.58} & \textbf{95.45 $\pm$ 0.62} & \textbf{99.87 $\pm$ 0.04} \\
    \bottomrule
    \end{tabular}%
    }
    \caption{Performance comparison of different models on the OrganCMNIST dataset. The table reports accuracy, F1, and AUC, with standard deviation when available. The best results are emphasized in bold.}
    \label{tab:performance_OrganCMNIST}
\end{table}

\begin{table}[h!]
    \centering
    \resizebox{0.6\textwidth}{!}{%
    \begin{tabular}{lccc}
    \toprule
    \textbf{Model} & \textbf{Accuracy (\%)} & \textbf{F1 (\%)} & \textbf{AUC (\%)} \\
    \midrule
    Resnet18~\cite{medmnistv2}& 77.8 & - &  97.4 \\
    Resnet50~\cite{medmnistv2}& 78.5 & - &  97.5 \\
    auto-sklearn~\cite{medmnistv2}& 67.2 & - &  94.5 \\
    AutoKeras~\cite{medmnistv2}& 81.3 & - &  97.4 \\
    Google AutoML Vision~\cite{medmnistv2}& 74.9 & - &  96.4 \\
    Resnet18~\cite{singh_dynamic_2024} & 81.3 & - & 97.4 \\
    Med ViT tiny~\cite{manzari_medvit_2023} & 78.9 & - & 97.2\\
    Med ViT small~\cite{manzari_medvit_2023} & 80.5 & - & \textbf{98.7}\\
    Med ViT base~\cite{manzari_medvit_2023} & 80.6 & - & 97.3\\\hline
    DINO small & 80.11 $\pm$ 1.51 & 74.66 $\pm$ 1.35 & 97.97 $\pm$ 0.26 \\
    DINO base & 82.44 $\pm$ 0.44 & 77.59 $\pm$ 0.27 & 97.21 $\pm$ 0.08 \\
    DINOv2 small & 77.87 $\pm$ 0.62 & 71.82 $\pm$ 0.80 & 97.62 $\pm$ 0.12 \\
    DINOv2 base & 73.97 $\pm$ 1.66 & 66.70 $\pm$ 2.17 & 97.01 $\pm$ 0.31 \\
    DINOv3 small & 83.66 $\pm$ 0.51 & 78.92 $\pm$ 0.86 & 98.16 $\pm$ 0.15 \\
    DINOv3 base & 82.65 $\pm$ 1.64 & 78.44 $\pm$ 1.27 & 98.00 $\pm$ 0.12 \\
    \hline
    Radio DINO small  & 82.3 $\pm$ 0.81 & 77.73 $\pm$ 1.2 & 98.27 $\pm$ 0.12 \\
    Radio DINO base &82.81 $\pm$ 0.96 & 78.15 $\pm$ 1.17 & 98.28 $\pm$ 0.15 \\
    \hline
    OmniRad small & 84.42 $\pm$ 0.33 & 79.91 $\pm$ 0.44 & 98.15 $\pm$ 0.14 \\
    OmniRad base & \textbf{85.28 $\pm$ 0.42} & \textbf{80.97 $\pm$ 0.45} & 98.33 $\pm$ 0.07 \\
    \bottomrule
    \end{tabular}%
    }
    \caption{Performance comparison of different models on the OrganSMNIST dataset. The table reports accuracy, F1, and AUC, with standard deviation when available. The best results are emphasized in bold.}
    \label{tab:performance_OrganSMNIST}
\end{table}

\subsection{Classification Ablation Study: Head-Only and LoRA Fine-Tuning}
\label{subsec:ablation_prompting}

To better understand the impact of different fine-tuning strategies on OmniRad base, we performed an ablation study across the MedMNIST classification benchmarks. We compare three configurations: full fine-tuning of all parameters (reported in~\Cref{tab:performance_BreastMNIST,tab:performance_PneumoniaMNIST,tab:performance_OrganAMNIST,tab:performance_OrganCMNIST,tab:performance_OrganSMNIST}), head-only fine-tuning with the backbone frozen, and LoRA adaptation with two parameter settings. Table~\ref{tab:ablation_prompting} reports the resulting F1 scores with standard deviations. Head-only fine-tuning consistently achieves strong performance, approaching the original full fine-tuning results, while LoRA~\cite{hu_lora_2021} variants provide slightly lower F1 scores. This demonstrates that even with a frozen backbone, careful adaptation can retain much of the model's capability, although full fine-tuning remains optimal for maximizing classification performance.

\begin{table}[h!]
    \centering
    \resizebox{0.6\textwidth}{!}{%
    \begin{tabular}{lcccc}
    \toprule
    \textbf{Dataset} & \textbf{Full} & \textbf{Head-Only} & \textbf{LoRA (R=8, $\alpha$=16)} & \textbf{LoRA (R=16, $\alpha$=32)} \\
    \midrule
    BreastMNIST      & \textbf{87.32 $\pm$ 3.72} & 86.54 $\pm$ 3.64 & 85.01 $\pm$ 3.68 & 84.44 $\pm$ 3.75 \\
    PneumoniaMNIST   & \textbf{94.15 $\pm$ 0.39} & 93.03 $\pm$ 0.48 & 91.51 $\pm$ 0.56 & 90.84 $\pm$ 0.61 \\
    OrganAMNIST      & \textbf{96.98 $\pm$ 0.55} & 96.03 $\pm$ 0.61 & 95.34 $\pm$ 0.63 & 94.72 $\pm$ 0.70 \\
    OrganCMNIST      & \textbf{95.45 $\pm$ 0.62} & 94.52 $\pm$ 0.68 & 93.62 $\pm$ 0.72 & 92.91 $\pm$ 0.78 \\
    OrganSMNIST      & \textbf{80.97 $\pm$ 0.45} & 79.81 $\pm$ 0.50 & 78.23 $\pm$ 0.55 & 77.52 $\pm$ 0.60 \\
    \bottomrule
    \end{tabular}%
    }
    \caption{Ablation study of OmniRad base on MedMNIST benchmarks. F1 scores with standard deviations are reported for full fine-tuning, head-only fine-tuning, and LoRA adaptation with two rank and scaling settings. Best scores in each row are highlighted in bold.}
    \label{tab:ablation_prompting}
\end{table}

\subsection{Segmentation Results}
\label{subsec:seg_results}
\begin{table*}[h!]
\centering
\resizebox{0.9\textwidth}{!}{%
\begin{tabular}{lcccc}
\toprule
\textbf{Model} & \textbf{BusiMSBench} & \textbf{CovidQUExMSBench} & \textbf{FHPsAOPMSBench} & \textbf{MosMedPlusMSBench} \\
\midrule
\multicolumn{5}{c}{\textbf{Baselines}} \\
\midrule
RN-18\cite{kus_medsegbench_2024}  & 57.80 & 62.70 & \textbf{92.90} & 67.40 \\
RN-50\cite{kus_medsegbench_2024}  & 54.70 & 62.00 & 92.30 & 68.20 \\
EN\cite{kus_medsegbench_2024}     & 62.40 & 63.30 & 92.70 & 67.40 \\
MN-v2\cite{kus_medsegbench_2024}  & 56.50 & 63.10 & 92.70 & 67.90 \\
DN-121\cite{kus_medsegbench_2024} & 61.50 & 64.70 & 92.80 & 68.60 \\
\midrule
\multicolumn{5}{c}{\textbf{Small models}} \\
\midrule

VIT small & 73.81 $\pm$ 0.76 & 82.57 $\pm$ 0.10 & 90.98 $\pm$ 0.06 & 84.53 $\pm$ 0.20 \\
DINO small & 73.14 $\pm$ 1.78 & 82.87 $\pm$ 0.22 & 91.01 $\pm$ 0.05 & 84.61 $\pm$ 0.27 \\
DINOv2 small & 75.23 $\pm$ 1.54 & 82.87 $\pm$ 0.10 & 91.16 $\pm$ 0.10 & 84.51 $\pm$ 0.14 \\
DINOv3 small& 71.84 $\pm$ 0.33 & 82.50 $\pm$ 0.17 & 91.13 $\pm$ 0.13 & 84.66 $\pm$ 0.15 \\
Radio DINO small & 75.79 $\pm$ 0.75 & 83.07 $\pm$ 0.13 & 91.32 $\pm$ 0.10 & 85.15 $\pm$ 0.10 \\
OmniRad small & 77.04 $\pm$ 0.41 & 83.19 $\pm$ 0.26 & 91.20 $\pm$ 0.05 & 84.87 $\pm$ 0.29 \\
\midrule
\multicolumn{5}{c}{\textbf{Base models}} \\
\midrule
VIT base& 75.62 $\pm$ 1.49 & 82.91 $\pm$ 0.25 & 91.40 $\pm$ 0.05 & 85.37 $\pm$ 0.20 \\
CLIP base& 70.64 $\pm$ 0.85 & 80.99 $\pm$ 0.52 & 91.26 $\pm$ 0.05 & 84.21 $\pm$ 0.32 \\
MAE base& 76.85 $\pm$ 1.15 & 83.61 $\pm$ 0.18 & 91.54 $\pm$ 0.15 & 85.62 $\pm$ 0.09 \\
DINO base& 77.18 $\pm$ 0.46 & 83.71 $\pm$ 0.14 & 91.55 $\pm$ 0.05 & 85.65 $\pm$ 0.24 \\
DINOv2 base& 75.24 $\pm$ 1.40 & 83.34 $\pm$ 0.17 & 91.52 $\pm$ 0.03 & 85.73 $\pm$ 0.07 \\
DINOv3 base& 68.68 $\pm$ 0.58 & 80.34 $\pm$ 0.43 & 91.16 $\pm$ 0.02 & 84.67 $\pm$ 0.25 \\
Radio DINO base& 76.57 $\pm$ 1.21 & 83.67 $\pm$ 0.40 & 91.52 $\pm$ 0.10 & 85.80 $\pm$ 0.21 \\
OmniRad base& \textbf{77.53 $\pm$ 1.36} & \textbf{83.77 $\pm$ 0.13} & 91.53 $\pm$ 0.05 & \textbf{86.19 $\pm$ 0.01} \\
\bottomrule
\end{tabular}
}
\caption{mIoU (\%) on Group 1 datasets}
\label{tab:seg_g1}

\end{table*}

\begin{table*}[h!]
\centering
\resizebox{0.9\textwidth}{!}{%

\begin{tabular}{lcccc}
\toprule
\textbf{Model} & \textbf{PandentalMSBench} & \textbf{Promise12MSBench} & \textbf{USforKidneyMSBench} & \textbf{UltrasoundNerveMSBench} \\
\midrule
\multicolumn{5}{c}{\textbf{Baselines}} \\
\midrule
RN-18\cite{kus_medsegbench_2024}  & 92.60 & 82.80 & 96.00 & 67.10 \\
RN-50\cite{kus_medsegbench_2024}  & 92.60 & 81.70 & 95.80 & 66.40 \\
EN\cite{kus_medsegbench_2024}     & 91.90 & 82.10 & 96.30 & 67.50 \\
MN-v2\cite{kus_medsegbench_2024}  & 90.70 & 82.70 & 96.10 & 66.00 \\
DN-121\cite{kus_medsegbench_2024} & \textbf{93.20} & 83.20 & 96.00 & 67.60 \\
\midrule
\multicolumn{5}{c}{\textbf{Small models}} \\
\midrule
VIT small & 91.28 $\pm$ 0.34 & 91.45 $\pm$ 0.19 & 96.96 $\pm$ 0.03 & 81.60 $\pm$ 0.11 \\
DINO small & 91.56 $\pm$ 0.36 & 91.17 $\pm$ 0.05 & 96.92 $\pm$ 0.03 & 81.29 $\pm$ 0.34 \\
DINOv2 small & 91.31 $\pm$ 0.43 & 91.97 $\pm$ 0.33 & 96.88 $\pm$ 0.01 & 81.33 $\pm$ 0.24 \\
DINOv3 small& 90.64 $\pm$ 0.33 & 91.61 $\pm$ 0.11 & 97.07 $\pm$ 0.04 & 81.53 $\pm$ 0.28 \\
Radio DINO small & 91.67 $\pm$ 0.31 & 92.50 $\pm$ 0.15 & 97.01 $\pm$ 0.01 & 81.85 $\pm$ 0.60 \\
OmniRad small & 91.16 $\pm$ 0.23 & 92.46 $\pm$ 0.19 & 96.95 $\pm$ 0.01 & 81.83 $\pm$ 0.13 \\
\midrule
\multicolumn{5}{c}{\textbf{Base models}} \\
\midrule
VIT base& 91.16 $\pm$ 0.41 & 91.81 $\pm$ 0.30 & 97.11 $\pm$ 0.09 & 82.30 $\pm$ 0.12 \\
CLIP base& 89.38 $\pm$ 0.61 & 90.09 $\pm$ 0.42 & 96.98 $\pm$ 0.02 & 81.64 $\pm$ 0.19 \\
MAE base& 92.86 $\pm$ 0.14 & 93.02 $\pm$ 0.08 & 97.16 $\pm$ 0.03 & 82.19 $\pm$ 0.23 \\
DINO base& 92.23 $\pm$ 0.09 & 92.70 $\pm$ 0.13 & 97.13 $\pm$ 0.04 & 82.19 $\pm$ 0.43 \\
DINOv2 base& 92.40 $\pm$ 0.57 & 93.02 $\pm$ 0.15 & 97.12 $\pm$ 0.05 & 82.20 $\pm$ 0.40 \\
DINOv3 base& 90.33 $\pm$ 0.48 & 91.54 $\pm$ 0.47 & 96.94 $\pm$ 0.06 & 81.92 $\pm$ 0.02 \\
Radio DINO base& 92.17 $\pm$ 0.20 & 92.80 $\pm$ 0.08 & 97.11 $\pm$ 0.02 & \textbf{82.61 $\pm$ 0.22} \\
OmniRad base& 91.66 $\pm$ 0.22 & \textbf{93.25 $\pm$ 0.45} & \textbf{97.25 $\pm$ 0.02} & 82.26 $\pm$ 0.15 \\
\bottomrule
\end{tabular}
}
\caption{mIoU (\%) on Group 2 datasets}
\label{tab:seg_g2}
\end{table*}

We evaluate the dense prediction capability of OmniRad, reporting performance in terms of mIoU, Dice coefficient, and F1, as summarized in \cref{tab:seg_g1,tab:seg_g2} and in the aggregated comparison in Table~\ref{tab:seg_all}.

\noindent\textbf{Overall performance.}

Across all benchmarks, OmniRad consistently achieves the strongest aggregated segmentation performance among all evaluated foundation models. In particular, the base variant attains the highest average mIoU $87.93\%$, Dice $92.95\%$, and F1 score $93.03\%$, establishing OmniRad as the optimal choice within this unified experimental setting. Compared to Radio DINO base, which represents the strongest competing baseline, OmniRad base yields systematic improvements across all major metrics, demonstrating that the proposed radiomics–oriented pretraining strategy produces more transferable and semantically coherent dense representations.

\noindent\textbf{Dataset–level behavior.}

OmniRad base achieves top performance across multiple benchmarks, including BusiMSBench, CovidQUExMSBench, and MosMedPlusMSBench, reaching a peak mIoU of $86.19\%$ on MosMedPlusMSBench. It also leads on Promise12MSBench and USforKidneyMSBench with $93.25\%$ and $97.25\%$ mIoU, respectively. These results span CT, MRI, and ultrasound, demonstrating that OmniRad captures modality–invariant radiological structures while preserving fine-grained anatomical boundaries.

\noindent\textbf{Model capacity effects and generalization}

Consistent with the trends observed in classification, the base variant generally outperforms the small model on large–scale and anatomically complex datasets, where increased representational capacity benefits dense anatomical delineation. Nevertheless, the small variant remains highly competitive and often surpasses existing foundation models, highlighting that OmniRad maintains strong transferability even in compact configurations.

The systematic improvements observed across all segmentation benchmarks confirm that OmniRad learns radiomics–aware visual representations that generalize effectively to dense prediction tasks across heterogeneous clinical imaging domains, supporting its role as a universal radiological foundation model for segmentation–driven clinical pipelines.

\begin{table}[h!]
\centering
\resizebox{0.55\textwidth}{!}{%
\begin{tabular}{lccc}
\toprule
\textbf{Model} & \textbf{mIoU (\%)} & \textbf{Dice (\%)} & \textbf{F1 (\%)} \\
\midrule
VIT small & 86.65 $\pm$ 7.05 & 92.03 $\pm$ 4.85 & 92.13 $\pm$ 4.73 \\
DINO small & 86.57 $\pm$ 7.23 & 91.99 $\pm$ 4.99 & 92.09 $\pm$ 4.84 \\
DINOv2 small & 86.91 $\pm$ 6.79 & 92.24 $\pm$ 4.61 & 92.33 $\pm$ 4.50 \\
DINOv3 small & 86.37 $\pm$ 7.54 & 91.84 $\pm$ 5.24 & 91.97 $\pm$ 5.06 \\
\midrule
VIT base& 87.21 $\pm$ 6.60 & 92.47 $\pm$ 4.42 & 92.55 $\pm$ 4.32 \\
DINO base& 87.79 $\pm$ 6.35 & 92.87 $\pm$ 4.17 & 92.96 $\pm$ 4.06 \\
MAE base& 87.86 $\pm$ 6.55 & 92.88 $\pm$ 4.30 & 92.97 $\pm$ 4.18 \\
DINOv2 base& 87.57 $\pm$ 6.89 & 92.65 $\pm$ 4.66 & 92.77 $\pm$ 4.48 \\
DINOv3 base& 85.70 $\pm$ 8.44 & 91.30 $\pm$ 6.06 & 91.49 $\pm$ 5.75 \\
CLIP base& 85.65 $\pm$ 7.74 & 91.36 $\pm$ 5.44 & 91.47 $\pm$ 5.31 \\
\midrule
Radio DINO small & 87.30 $\pm$ 6.68 & 92.54 $\pm$ 4.40 & 92.63 $\pm$ 4.29 \\
Radio DINO base& 87.78 $\pm$ 6.46 & 92.86 $\pm$ 4.23 & 92.93 $\pm$ 4.15 \\
\midrule
OmniRad small & 87.34 $\pm$ 6.34 & 92.57 $\pm$ 4.19 & 92.65 $\pm$ 4.11 \\
OmniRad base& \textbf{87.93 $\pm$ 6.30} & \textbf{92.95 $\pm$ 4.13} & \textbf{93.03 $\pm$ 4.03} \\
\bottomrule
\end{tabular}
}
\caption{Segmentation performance comparison. Mean $\pm$ standard deviation are reported for mIoU, Dice and F1. Best results are highlighted in bold.}
\label{tab:seg_all}
\end{table}

\begin{figure*}[htb!]
    \centering
    \includegraphics[width=1\linewidth]{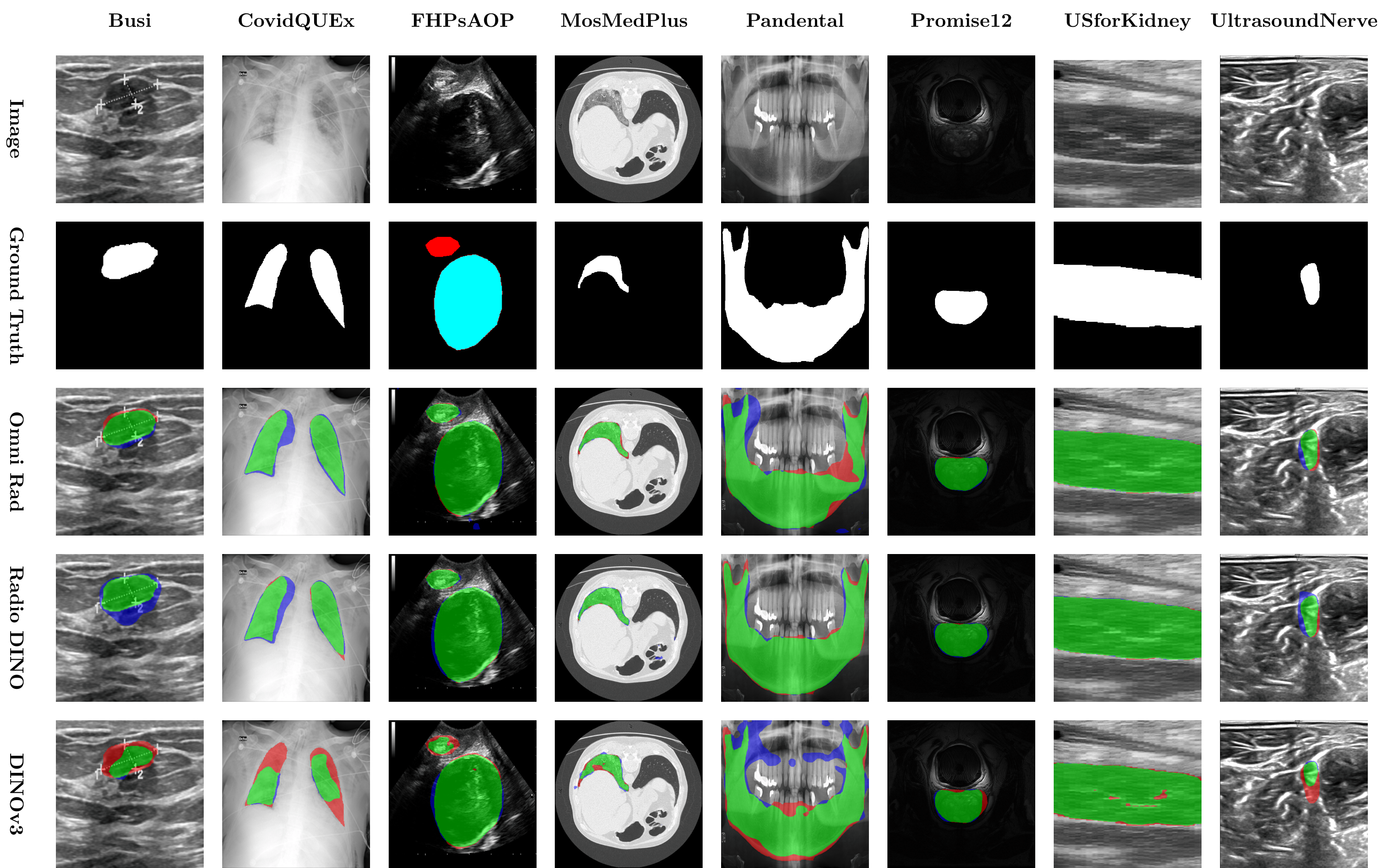}
    \caption{Segmentation visualization. Colors indicate prediction quality: green = correct prediction, blue = overprediction, red = missed prediction.}

    \label{fig:seg_qual}
\end{figure*}

\noindent\textbf{Qualitative results}  
Segmentation results highlight the strengths of OmniRad. Figure~\ref{fig:seg_qual} presents representative examples, showing that OmniRad often produces tighter predictions with minimal over- or under-segmentation, while also illustrating occasional failure cases, such as on the ParadentalMSBench dataset. Compared to general radiological foundation models, DINOv3 exhibits a higher rate of missed positives, emphasizing the benefits of specialized, radiology-focused models.

\subsection{Exploratory Captioning Results}
\label{subsec:captioning_results}

\begin{table*}[h!]
\centering
\small
\setlength{\tabcolsep}{4pt}
\resizebox{0.75\textwidth}{!}{%
\begin{tabular}{l c c r r r r r}
\toprule
\textbf{Image encoder} & \textbf{beams} & \textbf{max\_tokens} &
\textbf{METEOR} & \textbf{BLEU} & \textbf{BLEU-1} & \textbf{BLEU-4} & \textbf{ROUGE-L} \\
\midrule

CLIP base & 1 & 64 & 19.59 & 2.43 & 16.81 & 0.41 & 15.43 \\
CLIP base & 4 & 64 & 21.51 & 2.82 & 18.42 & 0.52 & 16.63 \\
CLIP base & 5 & 64 & 21.57 & 2.89 & 18.64 & 0.55 & 16.73 \\
CLIP base & 5 & 128 & 21.76 & 2.87 & 18.45 & \textbf{0.54} & 16.65 \\ \hline

DINOv2 base & 1 & 64 & 20.10 & 2.49 & 17.10 & 0.42 & 15.82 \\
DINOv2 base & 4 & 64 & 21.90 & 2.87 & 18.70 & 0.52 & 16.90 \\
DINOv2 base & 5 & 64 & 22.10 & 2.93 & 18.80 & \textbf{0.54} & 16.90 \\
DINOv2 base & 5 & 128 & 22.00 & 2.90 & 18.80 & 0.53 & 16.95 \\ \hline

DINOv3 base & 1 & 64 & 19.99 & 2.31 & 16.90 & 0.35 & 15.52 \\
DINOv3 base & 4 & 64 & 21.97 & 2.85 & 18.66 & 0.51 & 16.90 \\
DINOv3 base & 5 & 64 & 21.99 & 2.91 & 18.82 & \textbf{0.54} & 16.96 \\
DINOv3 base & 5 & 128 & 22.21 & 2.87 & 18.49 & 0.53 & 16.82 \\ \hline

Radio DINO base & 1 & 64 & 18.77 & 2.08 & 16.29 & 0.30 & 15.14 \\
Radio DINO base & 4 & 64 & 21.26 & 2.65 & 18.30 & 0.46 & 16.38 \\
Radio DINO base & 5 & 64 & 21.25 & 2.68 & 18.42 & 0.47 & 16.49 \\
Radio DINO base & 5 & 128 & 21.25 & 2.68 & 18.42 & 0.47 & 16.49 \\ \hline

OmniRad base & 1 & 64 & 20.42 & 2.57 & 17.40 & 0.44 & 15.98 \\
OmniRad base & 4 & 64 & 22.25 & 2.93 & 19.18 & 0.53 & 17.30 \\
OmniRad base & 5 & 64 & 22.33 & \textbf{2.97} & \textbf{19.39} & \textbf{0.54} & \textbf{17.48} \\
OmniRad base & 5 & 128 & \textbf{22.45} & 2.95 & 19.31 & 0.53 & 17.47 \\

\bottomrule
\end{tabular}}
\caption{Captioning benchmark.}
\label{tab:caption_clean}
\end{table*}

We perform an exploratory evaluation of OmniRad as a frozen visual encoder on the ROCOv2 benchmark using a unified BART–based decoding framework and a fixed experimental protocol across all encoders. Performance is reported in terms of METEOR, BLEU, BLEU–1, BLEU–4, and ROUGE–L, as summarized in Table~\ref{tab:caption_clean}.

\noindent\textbf{Overall performance.}

OmniRad base consistently achieves the strongest captioning performance across all evaluated metrics and decoding configurations. With beam size 5 and a maximum generation length of 64 tokens, OmniRad base reaches the highest BLEU score $2.97$, BLEU–1 $19.39$, BLEU–4 $0.54$, and ROUGE–L $17.48$, while also achieving the highest METEOR score $22.45$ when increasing the generation length to 128 tokens. These results indicate that OmniRad provides strong visual representations for radiological image captioning within this unified and controlled evaluation framework.

\noindent\textbf{Comparison with existing foundation models.}

Compared to CLIP, DINOv2, DINOv3, and Radio DINO, OmniRad consistently produces captions with higher automatic metric scores under the same decoding configuration, yielding systematic improvements across all major metrics. Notably, while DINOv2 and DINOv3 exhibit competitive performance under optimized decoding configurations, OmniRad maintains superior BLEU, ROUGE–L, and METEOR scores across all beam sizes, indicating a more stable and transferable representation for language grounding in radiological imagery.

\noindent\textbf{Effect of decoding configuration.}

Increasing the beam size from 1 to 5 leads to consistent improvements across all encoders, confirming the benefit of enhanced search during sequence generation. OmniRad shows particularly strong robustness to decoding variations, maintaining high captioning quality when increasing the maximum generation length, which further supports its ability to encode long–range anatomical and pathological semantics.

The consistent performance of OmniRad across decoding configurations suggests that its radiological representations support effective visual–semantic alignment in an exploratory captioning setting, complementing the gains observed in classification and segmentation tasks.

\begin{figure*}[!htb]
    \centering
    \includegraphics[width=1\linewidth]{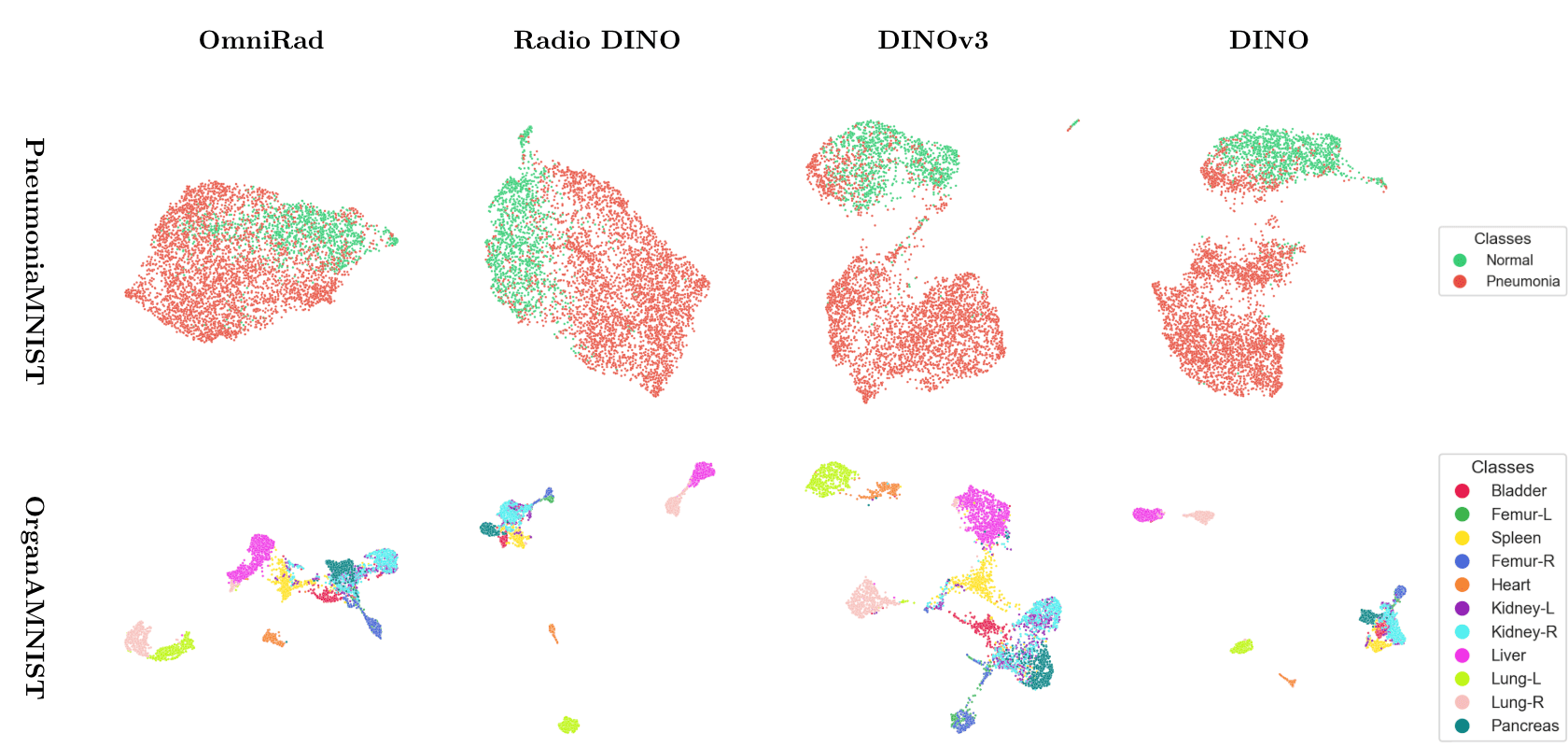}
    \caption{UMAP visualization of the latent representations learned by OmniRad, RadioDINO, and DINOv3. OmniRad exhibits reduced batch effects and a more structured embedding space.}
    \label{fig:umap}
\end{figure*}
\section{Discussion}
\label{sec:discussion}

The experimental results position OmniRad as a reliable pretrained visual backbone for a wide range of radiological tasks. While this study focuses on classification and segmentation, with an exploratory evaluation of image captioning, the consistency of the observed gains suggests that the learned representations may also benefit additional radiological applications not explicitly explored in this work.
Captioning results should therefore be interpreted as a proxy for visual-semantic alignment rather than a fully optimized report generation system.

A central contribution of OmniRad lies in demonstrating how architectural adaptations can address known limitations of plain transformer backbones in dense prediction settings. In particular, the proposed segmentation branch enables multi-scale feature extraction and allows OmniRad to remain competitive with, and often outperform, established U-Net–based architectures across heterogeneous segmentation benchmarks. These results indicate that transformer-based foundation models can be effectively adapted to radiologically dense tasks without sacrificing representation generality.

For image captioning, the evaluation is intentionally restricted to a controlled setting in which all visual encoders are paired with a shared decoder architecture. This design isolates the contribution of the image representations and enables a fair comparison across foundation models. While task-specific or fully fine-tuned architectures may further improve captioning performance, such extensions fall outside the scope of this work and would confound the assessment of representation quality.

Overall, OmniRad provides a stable backbone that supports multiple radiological tasks with minimal architecture-specific adaptation. This property reduces training complexity and computational cost, accelerating both research workflows and potential clinical deployment. When successfully integrated into real-world pipelines, such models may provide more consistent and reliable support for clinical decision-making.
From a clinical perspective, a single stable visual backbone reduces variability across tasks and studies, which is critical for reproducibility in longitudinal analysis and multi-center deployments.

\subsection{Statistical Analysis and Interpretation}
\label{subsec:stat_analysis}

To assess the robustness of OmniRad’s performance relative to existing baselines, we conducted a series of statistical comparisons across all classification and segmentation benchmarks. For datasets with multiple repeated runs, t-test was applied, whereas one-sample t-tests were used when the baseline consisted of a single value. 

Overall, the results indicate that OmniRad consistently matches or exceeds the performance of prior models, with statistically significant gains observed in datasets such as OrganSMNIST, MosMedPlusMSBench, and USforKidneyMSBench. These datasets are characterized by higher anatomical diversity or cross-modality variability, suggesting that OmniRad’s radiology-focused pretraining particularly benefits complex visual tasks. 

In several cases, including BreastMNIST, OrganAMNIST, PneumoniaMNIST, and Promise12MSBench, the performance differences are not statistically significant. Importantly, the absolute differences remain small, indicating that OmniRad is performing comparably to the strongest baselines even when significance is not reached. This tight performance range across benchmarks reflects both the stability of the learned representations and the low variance observed across repeated runs, which contributes to the consistency of the results. 

For segmentation datasets with single-sample baselines, such as FHPsAOPMSBench and PandentalMSBench, the t-tests indicate significant differences. In these cases, the baseline scores are slightly higher than OmniRad’s average, which is expected when comparing a model with repeated measurements to a fixed reference. Nevertheless, the absolute performance remains competitive, demonstrating that OmniRad produces high-quality dense predictions across multiple modalities, including CT, MRI, and ultrasound.

The frozen-encoder setting further emphasizes representation quality rather than task-specific optimization, aligning the evaluation with the intended use of OmniRad as a general-purpose radiological backbone.

\subsection{Latent space evaluation}

To complement the quantitative evaluation, we analyze the latent representations learned by the non–fine-tuned encoders using UMAP projections on PneumoniaMNIST and OrganAMNIST. As shown in Figure~\ref{fig:umap}, OmniRad exhibits reduced sensitivity to batch-related effects compared to DINO and DINOv3, whose embeddings display more fragmented clustering patterns.

In particular, OmniRad produces semantically coherent clusters, with anatomically related classes occupying nearby regions in the embedding space. For example, the left and right lung classes form adjacent clusters, suggesting a smooth representation of anatomical symmetry, while distinct structures such as the heart remain well separated. These qualitative observations indicate that OmniRad captures higher-level semantic relationships beyond class labels, which may contribute to its strong transfer performance across downstream tasks.

While UMAP visualizations provide only an approximate and qualitative view of the representation space, the observed structure is consistent with the improved robustness and generalization reported in the quantitative experiments.

This structured embedding behavior provides qualitative support for the low variance and consistent downstream performance observed across benchmarks.

\subsection{Limitations}
\label{subsec:limitations}

Despite consistent performance across classification, segmentation, and captioning pilots, several limitations should be considered.

First, OmniRad relies on RadImageNet initialization, which may not fully capture rare modalities, scanner-specific artifacts, or institution-dependent protocols. Residual domain biases may therefore persist, particularly in out-of-distribution settings. Extending pretraining to larger, multi-center clinical collections represents an important direction for future work.

Second, evaluation is limited to retrospective public benchmarks and does not include prospective clinical validation. Real-world diagnostic workflows often involve multi-view interpretation, longitudinal analysis, and integration of clinical metadata, which are not modeled in the present study.

Third, most experiments use a frozen-encoder paradigm to ensure computational efficiency and fair comparison. While this highlights representation quality, it may limit fine-grained task performance. Selective fine-tuning could address this limitation.

Finally, although OmniRad scales favorably from small to base configurations, training larger variants remains computationally demanding. Ultra-large models were not explored due to resource constraints, leaving scaling saturation and efficiency trade-offs unexamined. Captioning experiments are exploratory and evaluated using automatic metrics only, which do not fully capture clinical relevance, semantic accuracy, or safety concerns such as hallucinations. Human expert evaluation will be necessary to assess these aspects comprehensively.

These limitations do not negate the proposed approach but highlight directions for improving generalization, scalability, and clinical reliability.
Future work will focus on scaling pretraining to larger clinical cohorts and validating performance in prospective, institution-specific settings.
\section{Conclusions}
\label{sec:conclusions}

We introduced OmniRad, a radiology-driven foundation model designed to provide stable and transferable visual representations for diverse imaging tasks. Unlike task-specific or jointly optimized multi-task approaches, OmniRad follows a task-agnostic paradigm in which a single self-supervised encoder is reused across classification and segmentation, with an exploratory extension to image captioning. This design aligns with radiomics principles that emphasize representation stability, reproducibility, and reuse across clinical workflows.

Across a wide range of benchmarks spanning multiple modalities, anatomical regions, and task formulations, OmniRad demonstrated systematic improvements over existing medical foundation models. Gains were observed for image-level classification, dense anatomical segmentation, and report generation, confirming the versatility of the learned representations. Notably, these improvements were achieved with a frozen encoder during downstream adaptation, underscoring the robustness and generalizability of the pretrained features.

By combining large-scale self-supervised pretraining with a representation-focused adaptation strategy, OmniRad bridges classical radiomics and modern foundation models. Avoiding task-specific encoder retraining enables coherent feature reuse across tasks, which is particularly relevant for longitudinal studies, multi-center deployments, and integrated clinical pipelines where consistent representations are essential.
As such, OmniRad represents a practical step toward unified, reusable visual representations for radiology-driven foundation models in real-world clinical pipelines.

\section*{Code Availability}
The code for this study is available at the following GitHub repository: \url{https://github.com/unica-visual-intelligence-lab/OmniRad}. Additionally, we provide pretrained backbones on Hugging Face at \url{https://huggingface.co/collections/Snarcy/omnirad}.

\section*{Acknowledgement}
We acknowledge financial support under the National Recovery and Resilience Plan (NRRP), Mission 4 Component 2 Investment 1.5 - Call for tender No.3277 published on December 30, 2021 by the Italian Ministry of University and Research (MUR) funded by the European Union – NextGenerationEU. Project Code ECS0000038 – Project Title eINS Ecosystem of Innovation for Next Generation Sardinia – CUP F53C22000430001- Grant Assignment Decree No. 1056 adopted on June 23, 2022 by the Italian Ministry of University and Research (MUR).
\bibliographystyle{cas-model2-names}
\bibliography{bib} 

@String(CVPR  = {IEEE Conf. Comput. Vis. Pattern Recog.})

@String(ICCV  = {Int. Conf. Comput. Vis.})

@String(ICLR  = {Int. Conf. Learn. Represent.})

@String(CVPR  = {CVPR})

@String(ICCV  = {ICCV})

@String(ICLR  = {ICLR})

@article{oquab_dinov2_nodate,
  title={Dinov2: Learning robust visual features without supervision},
  author={Oquab, Maxime and Darcet, Timoth{\'e}e and Moutakanni, Th{\'e}o and Vo, Huy and Szafraniec, Marc and Khalidov, Vasil and Fernandez, Pierre and Haziza, Daniel and Massa, Francisco and El-Nouby, Alaaeldin and others},
  journal={arXiv preprint arXiv:2304.07193},
  year={2023}
}

@article{noauthor_radimagenet_nodate,
  title={RadImageNet: an open radiologic deep learning research dataset for effective transfer learning},
  author={Mei, Xueyan and Liu, Zelong and Robson, Philip M and Marinelli, Brett and Huang, Mingqian and Doshi, Amish and Jacobi, Adam and Cao, Chendi and Link, Katherine E and Yang, Thomas and others},
  journal={Radiology: Artificial Intelligence},
  volume={4},
  number={5},
  pages={e210315},
  year={2022},
  publisher={Radiological Society of North America}
}

@inproceedings{he_masked_2022,
	address = {New Orleans, LA, USA},
	title = {Masked {Autoencoders} {Are} {Scalable} {Vision} {Learners}},
	isbn = {978-1-66546-946-3},
	url = {https://ieeexplore.ieee.org/document/9879206/},
	doi = {10.1109/CVPR52688.2022.01553},
	language = {en},
	urldate = {2023-11-07},
	booktitle = {2022 {IEEE}/{CVF} {Conference} on {Computer} {Vision} and {Pattern} {Recognition} ({CVPR})},
	publisher = {IEEE},
	author = {He, Kaiming and Chen, Xinlei and Xie, Saining and Li, Yanghao and Dollar, Piotr and Girshick, Ross},
	month = jun,
	year = {2022},
	pages = {15979--15988}
}

@misc{koch_dinobloom_2024,
	title = {{DinoBloom}: {A} {Foundation} {Model} for {Generalizable} {Cell} {Embeddings} in {Hematology}},
	shorttitle = {{DinoBloom}},
	url = {http://arxiv.org/abs/2404.05022},
	doi = {10.48550/arXiv.2404.05022},
	urldate = {2024-07-23},
	publisher = {arXiv},
	author = {Koch, Valentin and Wagner, Sophia J. and Kazeminia, Salome and Sancar, Ece and Hehr, Matthias and Schnabel, Julia and Peng, Tingying and Marr, Carsten},
	month = apr,
	year = {2024},
	note = {arXiv:2404.05022 [cs]},
}

@article{medmnistv2,
    title={MedMNIST v2-A large-scale lightweight benchmark for 2D and 3D biomedical image classification},
    author={Yang, Jiancheng and Shi, Rui and Wei, Donglai and Liu, Zequan and Zhao, Lin and Ke, Bilian and Pfister, Hanspeter and Ni, Bingbing},
    journal={Scientific Data},
    volume={10},
    number={1},
    pages={41},
    year={2023},
    publisher={Nature Publishing Group UK London}
}

@InProceedings{Lai_2024_CVPR,
    author    = {Lai, Zhixin and Wu, Jing and Chen, Suiyao and Zhou, Yucheng and Hovakimyan, Naira},
    title     = {Residual-based Language Models are Free Boosters for Biomedical Imaging Tasks},
    booktitle = {Proceedings of the IEEE/CVF Conference on Computer Vision and Pattern Recognition (CVPR) Workshops},
    month     = {June},
    year      = {2024},
    pages     = {5086-5096}
}

@article{manzari_medvit_2023,
	title = {{MedViT}: {A} {Robust} {Vision} {Transformer} for {Generalized} {Medical} {Image} {Classification}},
	volume = {157},
	issn = {00104825},
	shorttitle = {{MedViT}},
	url = {http://arxiv.org/abs/2302.09462},
	doi = {10.1016/j.compbiomed.2023.106791},
	language = {en},
	urldate = {2024-09-01},
	journal = {Computers in Biology and Medicine},
	author = {Manzari, Omid Nejati and Ahmadabadi, Hamid and Kashiani, Hossein and Shokouhi, Shahriar B. and Ayatollahi, Ahmad},
	month = may,
	year = {2023},
	note = {arXiv:2302.09462 [cs]},
	pages = {106791},
}

@article{schafer_overcoming_2024,
	title = {Overcoming data scarcity in biomedical imaging with a foundational multi-task model},
	volume = {4},
	doi = {10.1038/s43588-024-00662-z},
	journal = {Nature Computational Science},
	author = {Schäfer, Raphael and Nicke, Till and Höfener, Henning and Lange, Annkristin and Merhof, Dorit and Feuerhake, Friedrich and Schulz, Volkmar and Lotz, Johannes and Kiessling, Fabian},
	month = jul,
	year = {2024},
	pages = {1--15},
}

@article{singh_dynamic_2024,
	title = {Dynamic {Filter} {Application} in {Graph} {Convolutional} {Networks} for {Enhanced} {Spectral} {Feature} {Analysis} and {Class} {Discrimination} in {Medical} {Imaging}},
	volume = {PP},
	doi = {10.1109/ACCESS.2024.3444042},
	journal = {IEEE Access},
	author = {Singh, Aryan and Ven, Pepijn and Eising, Ciarán and Denny, Patrick},
	month = jan,
	year = {2024},
	pages = {1--1},
}

@article{dosovitskiy_image_nodate,
  title={An Image is Worth 16x16 Words: Transformers for Image Recognition at Scale},
  author={Dosovitskiy, Alexey and Beyer, Lucas and Kolesnikov, Alexander and Weissenborn, Dirk and Zhai, Xiaohua and Unterthiner, Thomas and  Dehghani, Mostafa and Minderer, Matthias and Heigold, Georg and Gelly, Sylvain and Uszkoreit, Jakob and Houlsby, Neil},
  journal={ICLR},
  year={2021}
}

@article{chen2022vitadapter,
  title={Vision Transformer Adapter for Dense Predictions},
  author={Chen, Zhe and Duan, Yuchen and Wang, Wenhai and He, Junjun and Lu, Tong and Dai, Jifeng and Qiao, Yu},
  journal={arXiv preprint arXiv:2205.08534},
  year={2022}
}

@article{zedda_radio_2025,
    title = {Radio {DINO}: {A} foundation model for advanced radiomics and {AI}-driven medical imaging analysis},
    volume = {195},
    issn = {0010-4825},
    shorttitle = {Radio {DINO}},
    url = {https://www.sciencedirect.com/science/article/pii/S0010482525009345},
    doi = {10.1016/j.compbiomed.2025.110583},
    urldate = {2025-09-23},
    journal = {Computers in Biology and Medicine},
    author = {Zedda, Luca and Loddo, Andrea and Di Ruberto, Cecilia},
    month = sep,
    year = {2025},
    pages = {110583},
}

@misc{simeoni_dinov3_2025,
    title = {{DINOv3}},
    url = {http://arxiv.org/abs/2508.10104},
    doi = {10.48550/arXiv.2508.10104},
    urldate = {2025-09-23},
    publisher = {arXiv},
    author = {Siméoni, Oriane and Vo, Huy V. and Seitzer, Maximilian and Baldassarre, Federico and Oquab, Maxime and Jose, Cijo and Khalidov, Vasil and Szafraniec, Marc and Yi, Seungeun and Ramamonjisoa, Michaël and Massa, Francisco and Haziza, Daniel and Wehrstedt, Luca and Wang, Jianyuan and Darcet, Timothée and Moutakanni, Théo and Sentana, Leonel and Roberts, Claire and Vedaldi, Andrea and Tolan, Jamie and Brandt, John and Couprie, Camille and Mairal, Julien and Jégou, Hervé and Labatut, Patrick and Bojanowski, Piotr},
    month = aug,
    year = {2025},
    note = {arXiv:2508.10104 [cs]},
}

@article{isensee_nnu-net_2021,
    title = {{nnU}-{Net}: a self-configuring method for deep learning-based biomedical image segmentation},
    volume = {18},
    copyright = {2020 The Author(s), under exclusive licence to Springer Nature America, Inc.},
    issn = {1548-7105},
    shorttitle = {{nnU}-{Net}},
    url = {https://www.nature.com/articles/s41592-020-01008-z},
    doi = {10.1038/s41592-020-01008-z},
    language = {en},
    number = {2},
    urldate = {2026-01-13},
    journal = {Nature Methods},
    author = {Isensee, Fabian and Jaeger, Paul F. and Kohl, Simon A. A. and Petersen, Jens and Maier-Hein, Klaus H.},
    month = feb,
    year = {2021},
    note = {Publisher: Nature Publishing Group},
    pages = {203--211},
}

@misc{tian_yolov12_2025,
    title = {{YOLOv12}: {Attention}-{Centric} {Real}-{Time} {Object} {Detectors}},
    shorttitle = {{YOLOv12}},
    url = {http://arxiv.org/abs/2502.12524},
    doi = {10.48550/arXiv.2502.12524},
    urldate = {2026-01-13},
    publisher = {arXiv},
    author = {Tian, Yunjie and Ye, Qixiang and Doermann, David},
    month = feb,
    year = {2025},
    note = {arXiv:2502.12524 [cs]},
}

@misc{huang_real-time_2025,
    title = {Real-{Time} {Object} {Detection} {Meets} {DINOv3}},
    url = {http://arxiv.org/abs/2509.20787},
    doi = {10.48550/arXiv.2509.20787},
    urldate = {2026-01-13},
    publisher = {arXiv},
    author = {Huang, Shihua and Hou, Yongjie and Liu, Longfei and Yu, Xuanlong and Shen, Xi},
    month = sep,
    year = {2025},
    note = {arXiv:2509.20787 [cs]},
}

@inproceedings{raghu_vision_2021,
    title = {Do {Vision} {Transformers} {See} {Like} {Convolutional} {Neural} {Networks}?},
    volume = {34},
    url = {https://proceedings.neurips.cc/paper/2021/hash/652cf38361a209088302ba2b8b7f51e0-Abstract.html},
    urldate = {2026-01-14},
    booktitle = {Advances in {Neural} {Information} {Processing} {Systems}},
    publisher = {Curran Associates, Inc.},
    author = {Raghu, Maithra and Unterthiner, Thomas and Kornblith, Simon and Zhang, Chiyuan and Dosovitskiy, Alexey},
    year = {2021},
    pages = {12116--12128},
}

@misc{renggli_learning_2022,
    title = {Learning to {Merge} {Tokens} in {Vision} {Transformers}},
    url = {http://arxiv.org/abs/2202.12015},
    doi = {10.48550/arXiv.2202.12015},
    urldate = {2026-01-14},
    publisher = {arXiv},
    author = {Renggli, Cedric and Pinto, André Susano and Houlsby, Neil and Mustafa, Basil and Puigcerver, Joan and Riquelme, Carlos},
    month = feb,
    year = {2022},
    note = {arXiv:2202.12015 [cs]},
}

@inproceedings{lewis_bart_2020,
    address = {Online},
    title = {{BART}: {Denoising} {Sequence}-to-{Sequence} {Pre}-training for {Natural} {Language} {Generation}, {Translation}, and {Comprehension}},
    shorttitle = {{BART}},
    url = {https://aclanthology.org/2020.acl-main.703/},
    doi = {10.18653/v1/2020.acl-main.703},
    urldate = {2026-01-14},
    booktitle = {Proceedings of the 58th {Annual} {Meeting} of the {Association} for {Computational} {Linguistics}},
    publisher = {Association for Computational Linguistics},
    author = {Lewis, Mike and Liu, Yinhan and Goyal, Naman and Ghazvininejad, Marjan and Mohamed, Abdelrahman and Levy, Omer and Stoyanov, Veselin and Zettlemoyer, Luke},
    editor = {Jurafsky, Dan and Chai, Joyce and Schluter, Natalie and Tetreault, Joel},
    month = jul,
    year = {2020},
    pages = {7871--7880},
}

@article{ruckert_rocov2_2024,
    title = {{ROCOv2}: {Radiology} {Objects} in {COntext} {Version} 2, an {Updated} {Multimodal} {Image} {Dataset}},
    volume = {11},
    copyright = {2024 The Author(s)},
    issn = {2052-4463},
    shorttitle = {{ROCOv2}},
    url = {https://www.nature.com/articles/s41597-024-03496-6},
    doi = {10.1038/s41597-024-03496-6},
    language = {en},
    number = {1},
    urldate = {2026-01-14},
    journal = {Scientific Data},
    author = {Rückert, Johannes and Bloch, Louise and Brüngel, Raphael and Idrissi-Yaghir, Ahmad and Schäfer, Henning and Schmidt, Cynthia S. and Koitka, Sven and Pelka, Obioma and Abacha, Asma Ben and G. Seco de Herrera, Alba and Müller, Henning and Horn, Peter A. and Nensa, Felix and Friedrich, Christoph M.},
    month = jun,
    year = {2024},
    note = {Publisher: Nature Publishing Group},
    pages = {688},
}

@inproceedings{caron_emerging_2021,
    address = {Montreal, QC, Canada},
    title = {Emerging {Properties} in {Self}-{Supervised} {Vision} {Transformers}},
    isbn = {978-1-66542-812-5},
    url = {https://ieeexplore.ieee.org/document/9709990/},
    doi = {10.1109/ICCV48922.2021.00951},
    language = {en},
    urldate = {2023-11-07},
    booktitle = {2021 {IEEE}/{CVF} {International} {Conference} on {Computer} {Vision} ({ICCV})},
    publisher = {IEEE},
    author = {Caron, Mathilde and Touvron, Hugo and Misra, Ishan and Jegou, Herve and Mairal, Julien and Bojanowski, Piotr and Joulin, Armand},
    month = oct,
    year = {2021},
    pages = {9630--9640},
}

@article{koleilat_medclip-samv2_2025,
    title = {{MedCLIP}-{SAMv2}: {Towards} universal text-driven medical image segmentation},
    volume = {106},
    issn = {1361-8415},
    shorttitle = {{MedCLIP}-{SAMv2}},
    url = {https://www.sciencedirect.com/science/article/pii/S1361841525002968},
    doi = {10.1016/j.media.2025.103749},
    urldate = {2026-01-18},
    journal = {Medical Image Analysis},
    author = {Koleilat, Taha and Asgariandehkordi, Hojat and Rivaz, Hassan and Xiao, Yiming},
    month = dec,
    year = {2025},
    pages = {103749},
}

@article{lang_dacg_2025,
    title = {{DACG}: {Dual} {Attention} and {Context} {Guidance} model for radiology report generation},
    volume = {99},
    issn = {1361-8415},
    shorttitle = {{DACG}},
    url = {https://www.sciencedirect.com/science/article/pii/S1361841524003025},
    doi = {10.1016/j.media.2024.103377},
    urldate = {2026-01-18},
    journal = {Medical Image Analysis},
    author = {Lang, Wangyu and Liu, Zhi and Zhang, Yijia},
    month = jan,
    year = {2025},
    pages = {103377},
}

@article{gu_segmentanybone_2025,
    title = {{SegmentAnyBone}: {A} universal model that segments any bone at any location on {MRI}},
    volume = {101},
    issn = {1361-8415},
    shorttitle = {{SegmentAnyBone}},
    url = {https://www.sciencedirect.com/science/article/pii/S1361841525000179},
    doi = {10.1016/j.media.2025.103469},
    urldate = {2026-01-18},
    journal = {Medical Image Analysis},
    author = {Gu, Hanxue and Colglazier, Roy and Dong, Haoyu and Zhang, Jikai and Chen, Yaqian and Yildiz, Zafer and Chen, Yuwen and Li, Lin and Yang, Jichen and Willhite, Jay and Meyer, Alex M. and Guo, Brian and Shah, Yashvi Atul and Luo, Emily and Rajput, Shipra and Kuehn, Sally and Bulleit, Clark and Wu, Kevin A. and Lee, Jisoo and Ramirez, Brandon and Lu, Darui and Levin, Jay M. and Mazurowski, Maciej A.},
    month = apr,
    year = {2025},
    pages = {103469},
}

@article{yang_ddkg_2025,
    title = {{DDKG}: {A} {Dual} {Domain} {Knowledge} {Guidance} strategy for localization and diagnosis of non-displaced femoral neck fractures},
    volume = {100},
    issn = {1361-8415},
    shorttitle = {{DDKG}},
    url = {https://www.sciencedirect.com/science/article/pii/S1361841524003189},
    doi = {10.1016/j.media.2024.103393},
    urldate = {2026-01-18},
    journal = {Medical Image Analysis},
    author = {Yang, Jing and Wang, Lianxin and Lin, Chen and Wang, Jiacheng and Wang, Liansheng},
    month = feb,
    year = {2025},
    pages = {103393},
}

@article{park_self-supervised_2024,
    title = {Self-supervised multi-modal training from uncurated images and reports enables monitoring {AI} in radiology},
    volume = {91},
    issn = {1361-8415},
    url = {https://www.sciencedirect.com/science/article/pii/S1361841523002815},
    doi = {10.1016/j.media.2023.103021},
    urldate = {2026-01-19},
    journal = {Medical Image Analysis},
    author = {Park, Sangjoon and Lee, Eun Sun and Shin, Kyung Sook and Lee, Jeong Eun and Ye, Jong Chul},
    month = jan,
    year = {2024},
    pages = {103021},
}

@article{zhu_classification_2024,
    title = {Classification of lung cancer subtypes on {CT} images with synthetic pathological priors},
    volume = {95},
    issn = {1361-8415},
    url = {https://www.sciencedirect.com/science/article/pii/S1361841524001245},
    doi = {10.1016/j.media.2024.103199},
    urldate = {2026-01-19},
    journal = {Medical Image Analysis},
    author = {Zhu, Wentao and Jin, Yuan and Ma, Gege and Chen, Geng and Egger, Jan and Zhang, Shaoting and Metaxas, Dimitris N.},
    month = jul,
    year = {2024},
    pages = {103199},
}

@article{manigrasso_mammography_2025,
    title = {Mammography classification with multi-view deep learning techniques: {Investigating} graph and transformer-based architectures},
    volume = {99},
    issn = {1361-8415},
    shorttitle = {Mammography classification with multi-view deep learning techniques},
    url = {https://www.sciencedirect.com/science/article/pii/S1361841524002457},
    doi = {10.1016/j.media.2024.103320},
    urldate = {2026-01-19},
    journal = {Medical Image Analysis},
    author = {Manigrasso, Francesco and Milazzo, Rosario and Russo, Alessandro Sebastian and Lamberti, Fabrizio and Strand, Fredrik and Pagnani, Andrea and Morra, Lia},
    month = jan,
    year = {2025},
    pages = {103320},
}

@article{niu_medical_2025,
    title = {Medical multimodal multitask foundation model for lung cancer screening},
    volume = {16},
    copyright = {2025 The Author(s)},
    issn = {2041-1723},
    url = {https://www.nature.com/articles/s41467-025-56822-w},
    doi = {10.1038/s41467-025-56822-w},
    language = {en},
    number = {1},
    urldate = {2026-01-19},
    journal = {Nature Communications},
    author = {Niu, Chuang and Lyu, Qing and Carothers, Christopher D. and Kaviani, Parisa and Tan, Josh and Yan, Pingkun and Kalra, Mannudeep K. and Whitlow, Christopher T. and Wang, Ge},
    month = feb,
    year = {2025},
    note = {Publisher: Nature Publishing Group},
    pages = {1523},
}

@article{riberdy_radiomics_2025,
    title = {Radiomics in preclinical imaging research: methods, challenges and opportunities},
    volume = {3},
    copyright = {2025 The Author(s)},
    issn = {2948-197X},
    shorttitle = {Radiomics in preclinical imaging research},
    url = {https://www.nature.com/articles/s44303-025-00104-z},
    doi = {10.1038/s44303-025-00104-z},
    language = {en},
    number = {1},
    urldate = {2026-01-19},
    journal = {npj Imaging},
    author = {Riberdy, Vlora and Guida, Alessandro and Rioux, James and Brewer, Kimberly},
    month = sep,
    year = {2025},
    note = {Publisher: Nature Publishing Group},
    pages = {45},
}

@article{fernandez-miranda_retrospective_2024,
    title = {A retrospective study of deep learning generalization across two centers and multiple models of {X}-ray devices using {COVID}-19 chest-{X} rays},
    volume = {14},
    copyright = {2024 The Author(s)},
    issn = {2045-2322},
    url = {https://www.nature.com/articles/s41598-024-64941-5},
    doi = {10.1038/s41598-024-64941-5},
    language = {en},
    number = {1},
    urldate = {2026-01-19},
    journal = {Scientific Reports},
    author = {Fernández-Miranda, Pablo Menéndez and Fraguela, Enrique Marqués and de Linera-Alperi, Marta Alvarez and Cobo, Miriam and del Barrio, Amaia Pérez and González, David Rodríguez and Vega, José A. and Iglesias, Lara Lloret},
    month = jun,
    year = {2024},
    note = {Publisher: Nature Publishing Group},
    pages = {14657},
}

@article{wu_biologically_2024,
    title = {Biologically interpretable multi-task deep learning pipeline predicts molecular alterations, grade, and prognosis in glioma patients},
    volume = {8},
    copyright = {2024 The Author(s)},
    issn = {2397-768X},
    url = {https://www.nature.com/articles/s41698-024-00670-2},
    doi = {10.1038/s41698-024-00670-2},
    language = {en},
    number = {1},
    urldate = {2026-01-19},
    journal = {npj Precision Oncology},
    author = {Wu, Xuewei and Zhang, Shuaitong and Zhang, Zhenyu and He, Zicong and Xu, Zexin and Wang, Weiwei and Jin, Zhe and You, Jingjing and Guo, Yang and Zhang, Lu and Huang, Wenhui and Wang, Fei and Liu, Xianzhi and Yan, Dongming and Cheng, Jingliang and Yan, Jing and Zhang, Shuixing and Zhang, Bin},
    month = aug,
    year = {2024},
    note = {Publisher: Nature Publishing Group},
    pages = {181},
}

@article{zhao_multi-task_2023,
    title = {Multi-task deep learning for medical image computing and analysis: {A} review},
    volume = {153},
    issn = {0010-4825},
    shorttitle = {Multi-task deep learning for medical image computing and analysis},
    url = {https://www.sciencedirect.com/science/article/pii/S0010482522012045},
    doi = {10.1016/j.compbiomed.2022.106496},
    urldate = {2026-01-19},
    journal = {Computers in Biology and Medicine},
    author = {Zhao, Yan and Wang, Xiuying and Che, Tongtong and Bao, Guoqing and Li, Shuyu},
    month = feb,
    year = {2023},
    pages = {106496},
}

@article{yao_deep_2025,
    title = {Deep {Learning} {Applications} in {Clinical} {Cancer} {Detection}: {A} {Review} of {Implementation} {Challenges} and {Solutions}},
    volume = {3},
    issn = {2949-7612},
    shorttitle = {Deep {Learning} {Applications} in {Clinical} {Cancer} {Detection}},
    url = {https://www.sciencedirect.com/science/article/pii/S2949761225000604},
    doi = {10.1016/j.mcpdig.2025.100253},
    number = {3},
    urldate = {2026-01-19},
    journal = {Mayo Clinic Proceedings: Digital Health},
    author = {Yao, Isaiah Z. and Dong, Min and Hwang, William Y. K.},
    month = sep,
    year = {2025},
    pages = {100253},
}

@article{shen_multi-modal_2024,
    title = {Multi-modal large language models in radiology: principles, applications, and potential},
    volume = {50},
    copyright = {2024 The Author(s), under exclusive licence to Springer Science+Business Media, LLC, part of Springer Nature},
    issn = {2366-0058},
    shorttitle = {Multi-modal large language models in radiology},
    url = {https://link.springer.com/article/10.1007/s00261-024-04708-8},
    doi = {10.1007/s00261-024-04708-8},
    language = {En},
    number = {6},
    urldate = {2026-01-19},
    journal = {Abdominal Radiology},
    author = {Shen, Yiqiu and Xu, Yanqi and Ma, Jiajian and Rui, Wushuang and Zhao, Chen and Heacock, Laura and Huang, Chenchan},
    month = dec,
    year = {2024},
    note = {Publisher: Springer},
    pages = {2745--2757},
}

@article{kus_medsegbench_2024,
    title = {{MedSegBench}: {A} comprehensive benchmark for medical image segmentation in diverse data modalities},
    volume = {11},
    copyright = {2024 The Author(s)},
    issn = {2052-4463},
    shorttitle = {{MedSegBench}},
    url = {https://www.nature.com/articles/s41597-024-04159-2},
    doi = {10.1038/s41597-024-04159-2},
    language = {en},
    number = {1},
    urldate = {2025-02-07},
    journal = {Scientific Data},
    author = {Kuş, Zeki and Aydin, Musa},
    month = nov,
    year = {2024},
    note = {Publisher: Nature Publishing Group},
    pages = {1283},
}

@article{lambin_radiomics_2012,
    title = {Radiomics: {Extracting} more information from medical images using advanced feature analysis},
    volume = {48},
    issn = {0959-8049},
    shorttitle = {Radiomics},
    url = {https://www.sciencedirect.com/science/article/pii/S0959804911009993},
    doi = {10.1016/j.ejca.2011.11.036},
    number = {4},
    urldate = {2026-01-19},
    journal = {European Journal of Cancer},
    author = {Lambin, Philippe and Rios-Velazquez, Emmanuel and Leijenaar, Ralph and Carvalho, Sara and van Stiphout, Ruud G. P. M. and Granton, Patrick and Zegers, Catharina M. L. and Gillies, Robert and Boellard, Ronald and Dekker, André and Aerts, Hugo J. W. L.},
    month = mar,
    year = {2012},
    pages = {441--446},
}

@article{zhong_multi-modal_2026,
    title = {Multi-modal multi-scale representation learning via cross-attention between chest radiology images and free-text reports},
    volume = {111},
    issn = {1746-8094},
    url = {https://www.sciencedirect.com/science/article/pii/S1746809425008298},
    doi = {10.1016/j.bspc.2025.108318},
    urldate = {2026-01-19},
    journal = {Biomedical Signal Processing and Control},
    author = {Zhong, Daidi and Li, Xiaoyu and Huang, Zhiyong and Wang, Shiwei and Yu, Zhi and Hou, Mingyang and Yan, Yan and Liu, Yushi},
    month = jan,
    year = {2026},
    pages = {108318},
}

@article{dantonoli_foundation_nodate,
    title = {Foundation models for radiology: fundamentals, applications, opportunities, challenges, risks, and prospects},
    shorttitle = {Foundation models for radiology},
    journal = {Diagn Interv Radiol.},
    url = {https://www.dirjournal.org/articles/foundation-models-for-radiology-fundamentals-applications-opportunities-challenges-risks-and-prospects/doi/dir.2025.253445},
    doi = {10.4274/dir.2025.253445},
    urldate = {2026-01-19},
    author = {D’Antonoli, Tugba Akinci and Bluethgen, Christian and Cuocolo, Renato and Klontzas, Michail E. and Ponsiglione, Andrea and Kocak, Burak and D’Antonoli, Tugba Akinci and Bluethgen, Christian and Cuocolo, Renato and Klontzas, Michail E. and Ponsiglione, Andrea and Kocak, Burak},
    note = {Publisher: Diagnostic and Interventional Radiology},
    year = {2025}
}

@article{zhong_abn-blip_2026,
    title = {Abn-{BLIP}: {Abnormality}-aligned {Bootstrapping} {Language}-{Image} {Pre}-training for pulmonary embolism diagnosis and report generation from {CTPA}},
    volume = {107},
    issn = {1361-8415},
    shorttitle = {Abn-{BLIP}},
    url = {https://www.sciencedirect.com/science/article/pii/S1361841525003329},
    doi = {10.1016/j.media.2025.103786},
    urldate = {2026-01-19},
    journal = {Medical Image Analysis},
    author = {Zhong, Zhusi and Wang, Yuli and Bi, Lulu and Ma, Zhuoqi and Ahn, Sun Ho and Mullin, Christopher J. and Greineder, Colin F. and Atalay, Michael K. and Collins, Scott and Baird, Grayson L. and Lin, Cheng Ting and Stayman, J. Webster and Kolb, Todd M. and Kamel, Ihab and Bai, Harrison X. and Jiao, Zhicheng},
    month = jan,
    year = {2026},
    pages = {103786},
}

@article{okolo_cln_2025,
    title = {{CLN}: {A} multi-task deep neural network for chest {X}-ray image localisation and classification},
    volume = {288},
    issn = {0957-4174},
    shorttitle = {{CLN}},
    url = {https://www.sciencedirect.com/science/article/pii/S0957417425017828},
    doi = {10.1016/j.eswa.2025.128162},
    urldate = {2026-01-19},
    journal = {Expert Systems with Applications},
    author = {Okolo, Gabriel Iluebe and Katsigiannis, Stamos and Ramzan, Naeem},
    month = sep,
    year = {2025},
    pages = {128162},
}

@article{kim_communication_2025,
    title = {Communication {Efficient} {Federated} {Learning} for {Multi}-{Organ} {Segmentation} via {Knowledge} {Distillation} {With} {Image} {Synthesis}},
    volume = {44},
    issn = {1558-254X},
    url = {https://ieeexplore.ieee.org/document/10829700},
    doi = {10.1109/TMI.2025.3525581},
    number = {5},
    urldate = {2026-01-19},
    journal = {IEEE Transactions on Medical Imaging},
    author = {Kim, Soopil and Park, Heejung and Chikontwe, Philip and Kang, Myeongkyun and Hwan Jin, Kyong and Adeli, Ehsan and Pohl, Kilian M. and Hyun Park, Sang},
    month = may,
    year = {2025},
    pages = {2079--2092},
}

@misc{hu_lora_2021,
    title = {{LoRA}: {Low}-{Rank} {Adaptation} of {Large} {Language} {Models}},
    shorttitle = {{LoRA}},
    url = {http://arxiv.org/abs/2106.09685},
    doi = {10.48550/arXiv.2106.09685},
    urldate = {2025-06-10},
    publisher = {arXiv},
    author = {Hu, Edward J. and Shen, Yelong and Wallis, Phillip and Allen-Zhu, Zeyuan and Li, Yuanzhi and Wang, Shean and Wang, Lu and Chen, Weizhu},
    month = oct,
    year = {2021},
    note = {arXiv:2106.09685 [cs]},
}
\end{document}